\documentclass{article} 
\usepackage{iclr2027_conference,times}

\usepackage{amsmath,amsfonts,bm}

\def\eqref#1{equation~\ref{#1}}

\def\1{\bm{1}}

\DeclareMathAlphabet{\mathsfit}{\encodingdefault}{\sfdefault}{m}{sl}
\SetMathAlphabet{\mathsfit}{bold}{\encodingdefault}{\sfdefault}{bx}{n}

\usepackage{hyperref}

\usepackage{url}
\usepackage{amsmath,amssymb,amsthm}
\usepackage{mathtools}
\usepackage{bm}
\usepackage{xcolor}
\usepackage{hyperref}
\usepackage{cleveref}
\usepackage{natbib}
\usepackage{multirow}
\usepackage{graphicx}
\usepackage{booktabs}
\usepackage{diagbox}
\usepackage{wrapfig}
\usepackage{adjustbox}
\usepackage[table]{xcolor}  
\usepackage{soul}           
\usepackage{array}
\usepackage{booktabs}
\usepackage{subcaption}
\usepackage{listings}
\usepackage{tcolorbox}
\tcbuselibrary{skins,breakable,listings}
\usepackage{xcolor}
\usepackage{pifont} 
\usepackage{subcaption}
\usepackage{algorithm}
\usepackage{algpseudocode}
\usepackage{enumitem}
\usepackage{caption}
\usepackage{subcaption}
\usepackage{url}
\usepackage{textcomp}
\usepackage{arydshln}
\usepackage{cite}

\usepackage{wrapfig}

\usepackage{xcolor}
\usepackage{listings}

\definecolor{PromptBlue}{HTML}{9CC9E8}
\definecolor{PromptTitleBlue}{HTML}{EAF4FC}

\newtcblisting{specbox}[1]{
    enhanced,
    breakable,
    sharp corners,
    title={#1},
    colframe=PromptBlue,
    colback=white,
    colbacktitle=PromptTitleBlue,
    coltitle=black,
    fonttitle=\small\bfseries,
    boxrule=1pt,
    boxsep=1pt,
    left=5pt,
    right=5pt,
    top=5pt,
    bottom=5pt,
    before skip=10pt,
    after skip=10pt,
    listing only,
    listing options={
        language={},
        basicstyle=\footnotesize\ttfamily,
        breaklines=true,
        breakatwhitespace=false,
        columns=fullflexible,
        keepspaces=true,
        showstringspaces=false,
        showspaces=false,
        showtabs=false,
        tabsize=2,
        aboveskip=0pt,
        belowskip=0pt
    }
}

\definecolor{light-green}{RGB}{220,255,220}
\definecolor{light-blue}{cmyk}{0.1,0,0,0}

\newcommand{\circled}[1]{\ding{\numexpr171+#1\relax}} 

\title{All-in-One Memory: Process-Rewarded Reinforcement Learning for Memory Agents}

\author{ Bingbing Wang$^{1,2*}$, Jingjie Lin$^{1*}$,Weiming Qiao$^{3*}$, Jing Li$^{2}$, Ruifeng Xu$^{1,4\dag}$ \\
$^{1}$ Harbin Institute of Technology, Shenzhen, $^{2}$ The Hong Kong Polytechnic University, \\  $^{3}$ Fudan University
, $^{4}$ Peng Cheng Laboratory, bingbing.wang@stu.hit.edu.cn}

\iclrfinalcopy
\begin{document}

\maketitle

\begin{abstract}
Large language models increasingly rely on external memory to answer queries grounded in long-horizon histories such as multi-session dialogues and lengthy documents. However, existing memory methods often optimize retrieval and memory construction separately, without a shared reward space for evaluating search, writing, and stopping decisions. To address this, we propose \textbf{A}ll-\textbf{i}n-\textbf{O}ne \textbf{Mem}ory (\textbf{AiO-Mem}), which organizes these operations as transitions over a shared evidence state and directly evaluates the semantic quality of each action within a unified reward space. AiO-Mem compiles heterogeneous histories into source-traceable multi-view events, providing a shared substrate for retrieval and memory writing. A unified memory policy actor then rolls out complete interaction trajectories over this shared evidence state. A separate trajectory-conditioned process reward model is trained through evidence-grounded supervised fine-tuning and trajectory-level DPO to assign process scores to the collected trajectories. These scores are aggregated into an action-balanced trajectory reward and used by Process-Aligned DAPO (PA-DAPO) to optimize the memory policy. Across long-context memory benchmarks, AiO-Mem consistently improves answer accuracy over both static memory systems and RL-trained memory agents.
\end{abstract}

\section{Introduction}


Large language models (LLMs) have achieved remarkable progress across diverse natural language processing benchmarks \citep{openai2023gpt4, touvron2023llama}. Yet in long-context, multi-step reasoning scenarios, their static parametric knowledge and limited context windows remain critical bottlenecks \citep{liu2024lost}. External memory-augmented architectures have therefore become important extensions of LLMs, ranging from dense retrieval systems \citep{lewis2020rag, guu2020realm} and knowledge-graph memories \citep{edge2024graphrag} to persistent memory agents \citep{packer2023memgpt, chhikara2025mem0}. Such systems rely on tightly coupled memory retrieval and memory construction: retrieval supplies relevant evidence, while construction converts it into compact working memory. Coordinating these stages is challenging because \textbf{the evolving memory state continually changes what evidence should be searched for, written, or discarded}.

\begin{figure}[!t]
    \centering
    \vspace{0em}
    \includegraphics[width=0.85\linewidth]{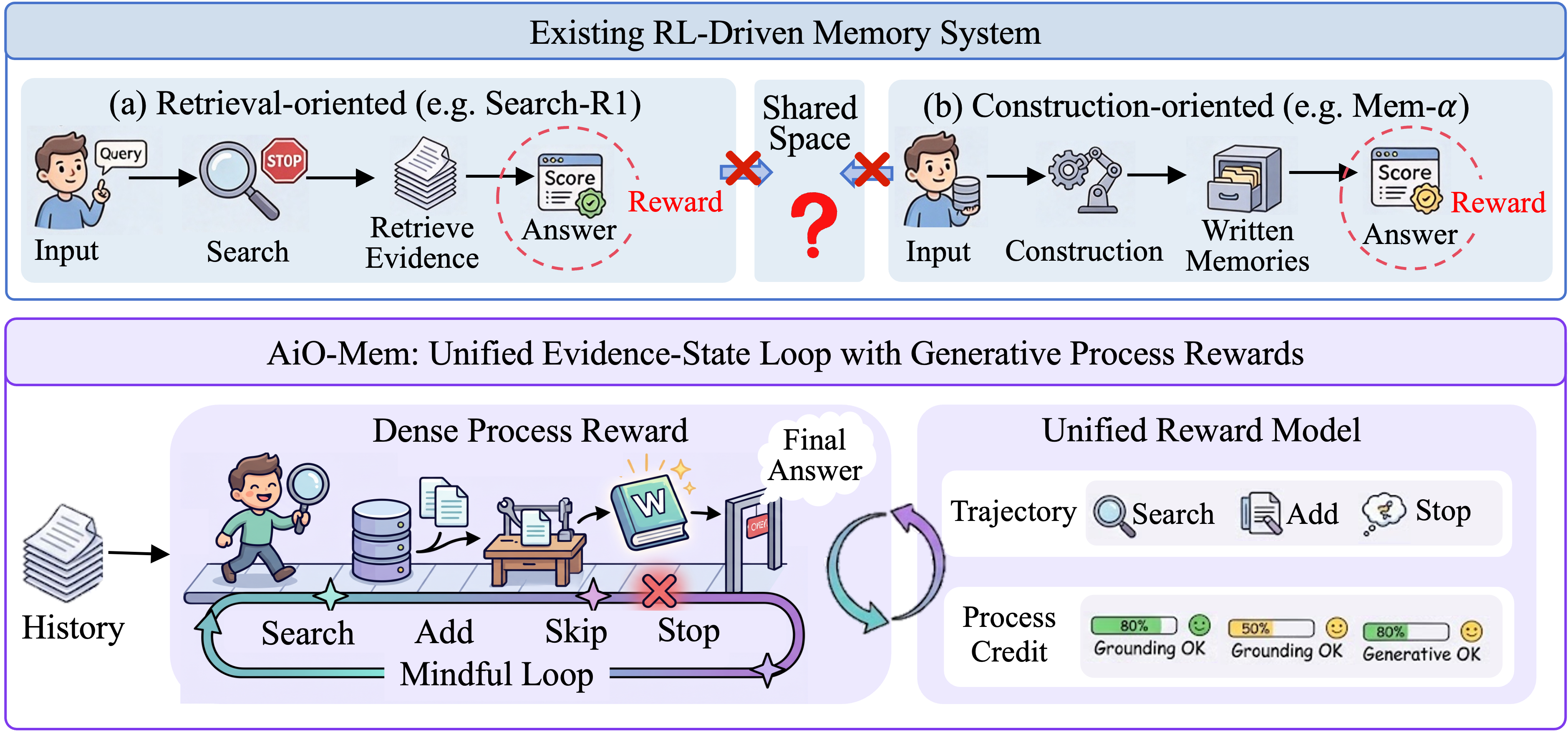}
    \caption{Comparison of existing RL-driven memory methods and AiO-Mem.}
    \label{fig:comparison}
    \vspace{-1em}
\end{figure}

Recent studies have introduced reinforcement learning (RL) into memory systems to enable self-improving memory behavior. As illustrated in Figure~\ref{fig:comparison}, existing RL-driven memory systems are typically optimized along two separate routes: retrieval-oriented methods, such as Search-R1 \citep{jin2025searchr1}, learn how to search for evidence, while construction-oriented methods, such as Mem-$\alpha$ \citep{wang2025memalpha}, learn when to insert, update, or merge memory entries. However, these methods still optimize retrieval and construction in isolation, leaving the two stages governed by separate objectives and evaluated through partially disconnected feedback signals. This separation exposes two key limitations. First, there is no shared evidence-state space in which retrieved evidence, written memories, and stopping decisions can be jointly evaluated, making it difficult to coordinate whether the system should continue searching, write a memory, skip redundant information, or stop. Second, existing supervision is largely outcome-centered: it assigns rewards to the final answer but provides little process-level credit for the intermediate memory operations that produced it. As a result, trajectories with the same final-answer score may still differ substantially in evidence relevance, memory faithfulness, redundancy control, and stopping sufficiency; \textbf{answer correctness alone cannot reliably evaluate the semantic quality of memory operations.}


To overcome these limitations, we propose \textbf{A}ll-\textbf{i}n-\textbf{O}ne \textbf{Mem}ory (\textbf{AiO-Mem}), which unifies memory retrieval and construction as evidence-state transitions of a single memory-interaction agent, and uses a separate trajectory-conditioned reward model to evaluate the resulting trajectories. AiO-Mem first compiles heterogeneous histories into a traceable multi-view event bank, providing a shared source-grounded substrate for retrieval and memory writing. Built on this event bank, the memory policy actor operates over a shared evidence state and rolls out complete interaction trajectories through two sequential decision steps: \textit{decide}, which determines whether to continue searching or stop, and \textit{extract}, which determines whether retrieved candidates should be written into working memory or discarded. A unified process reward model is then trained through evidence-grounded SFT and trajectory-level DPO to assign process scores to each action in the collected trajectories, using rule-derived supervision and reasoning-model scores as complementary training signals. Finally, these process scores are aggregated into an action-balanced trajectory reward. Building on DAPO \citep{yu2026dapo}, we propose \underline{P}rocess-\underline{A}ligned DAPO (\underline{PA}-DAPO) to optimize the memory policy actor. We summarize our primary contributions as follows:
\begin{itemize}[leftmargin=1.2em, itemsep=2pt]
    \item We propose \textbf{AiO-Mem}, an all-in-one memory agent that unifies retrieval, writing, filtering, and termination within a single evidence-state transition loop. This formulation turns memory retrieval and construction from separately optimized modules into one jointly trainable policy over source-traceable evidence.
    \item We introduce a \textbf{trajectory-conditioned unified reward model} that directly assesses the semantic quality of each memory action within a shared evidence state, scoring search, add, skip, and stop in a common reward space. Action-balanced aggregation then converts these scores into a trajectory reward for memory-policy optimization with our proposed PA-DAPO.
    \item We conduct extensive experiments on long-context memory benchmarks, where AiO-Mem consistently improves both answer accuracy and reasoning faithfulness over static memory systems and RL-trained memory agents. 
\end{itemize}

\section{Related Work}

\textbf{Memory-Augmented Large Language Models.}
External memory equips LLMs with non-parametric knowledge beyond their context windows. Early retrieval-augmented models, such as RAG~\citep{lewis2020rag} and REALM~\citep{guu2020realm} established dense retrieval as a standard way to ground generation in external evidence. Later systems enrich retrieval with structured or task-specific memory organization: GraphRAG~\citep{edge2024graphrag}, HippoRAG~\citep{gutierrez2024hipporag}, KG-RAG~\citep{sanmartin2024kgrag}, and Emotional RAG~\citep{huang2024emotionalrag} incorporate entity graphs, graph traversal, knowledge-graph grounding, or affective similarity.
A complementary line studies persistent memory across interactions. MemGPT~\citep{packer2023memgpt} introduces virtual memory tiers; MemoryBank~\citep{zhong2024memorybank} and Mem0~\citep{chhikara2025mem0} maintain user-centric summaries; MemoryOS~\citep{kang2025memoryos} and LightMem~\citep{fang2026lightmem} organize memories into tiered or compressed stores; and A-Mem~\citep{xu2026amem}, Zep~\citep{rasmussen2025zep}, EverMemOS~\citep{hu2026evermemos}, LangMem~\citep{langchain2025langmem}, and GAM~\citep{yan2025gam} explore atomic notes, MemCell structures, online consolidation, and iterative evidence gathering.
These systems provide increasingly rich retrieval and construction mechanisms, but most coordinate memory operations through predefined workflows, hand-designed heuristics, or modular pipelines rather than a learned policy that jointly optimizes retrieval, writing, filtering, and termination.

\textbf{Reinforcement Learning for Memory Policies.}
Recent work has applied RL to memory systems at different stages of the pipeline. Retrieval-oriented methods, including Search-R1~\citep{jin2025searchr1}, R1-Searcher~\citep{song2025r1searcher}, and DeepRetrieval~\citep{jiang2025deepretrieval}, learn search or retrieval strategies that improve downstream answers. Persistent-memory methods, such as Memory-R1~\citep{yan2026memoryr1}, Mem-$\alpha$~\citep{wang2025memalpha}, and AtomMem~\citep{huo2026atommem}, optimize memory insertion, update, merging, or atomic memory management. More recent studies, including Fine-Mem~\citep{ma2026finemem}, TRUSTMEM~\citep{yang2026trustmem}, and Mem-T~\citep{yue2026memt}), move toward process-level supervision for memory use. Despite this progress, most RL-based memory methods either rely primarily on final-answer correctness or provide process feedback within a restricted stage of the memory pipeline. As a result, they do not directly evaluate heterogeneous memory actions within a shared evidence state and a common reward space. AiO-Mem addresses this gap by learning a trajectory-conditioned reward model that assigns process-level semantic scores to different memory actions and uses them to optimize a unified memory policy.

\begin{figure}[!t]
  \centering
\includegraphics[width=\linewidth]{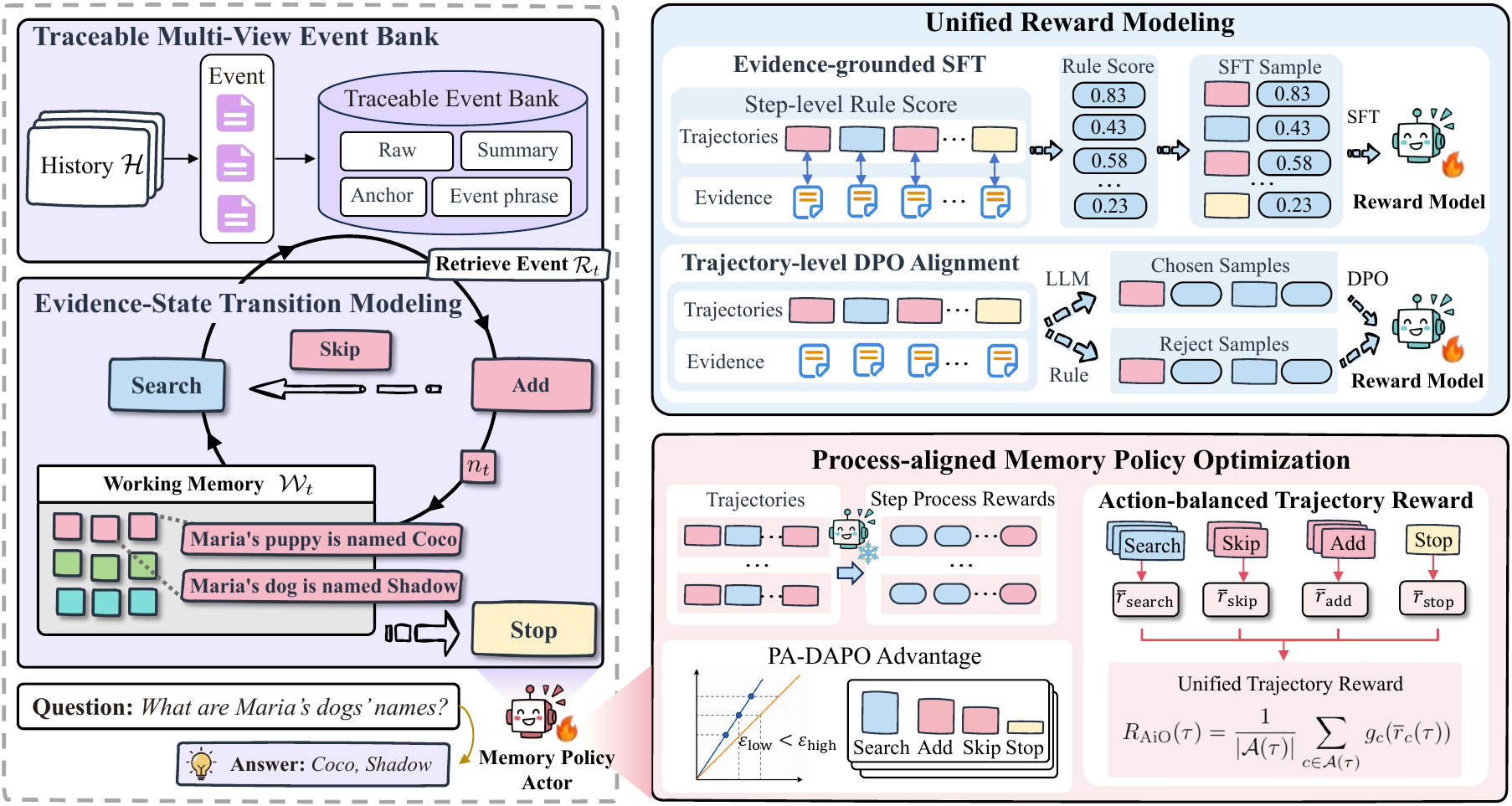}
  \caption{Overview of our AiO-Mem method.
}
  \label{fig:method}
  \vspace{-8pt}
\end{figure}

\section{Method}
\label{sec:method}

\subsection{Problem Formulation and Method Overview}
\label{sec:formulation}

\textbf{Problem Formulation.} We study memory-augmented question answering (QA) over long-horizon histories. 
Let $q$ be a query and $\mathcal{H}=(h_1,\ldots,h_N)$ be its associated history of $N$ heterogeneous source units, such as dialogue sessions, document passages, or facts. 
Given $(q,\mathcal{H})$, a memory policy constructs a compact working memory $\mathcal{W}$ from which a frozen downstream generator $\hat{y}=\mathcal{G}(q,\mathcal{W})$. 
The learning objective is to acquire an optimal sequence of memory operations that transforms a dispersed history into a reasoning-sufficient evidence state.

\textbf{Method Overview.} As shown in Figure~\ref{fig:method}, we propose \textbf{AiO-Mem}, a four-component framework that trains a memory policy actor $\pi_\theta$ to construct working memory through sequential evidence-state transitions. AiO-Mem integrates four components for memory construction, interaction modeling, process supervision, and policy optimization:
(i) \emph{Traceable Multi-View Event Bank} (Section~\ref{sec:event-memory}), which compiles $\mathcal{H}$ into source-traceable event units and provides the retrieval substrate for memory interaction;
(ii) \emph{Evidence-State Transition Modeling} (Section~\ref{sec:memory-interaction}), which formulates the interaction of $\pi_\theta$ as a finite-horizon MDP over a shared evidence-state space, with four action types organized into a \emph{decide} step (\textbf{search} or \textbf{stop}) and, after each search, an \emph{extract} step (\textbf{add} or \textbf{skip});
(iii) \emph{Unified Reward Modeling} (Section~\ref{sec:reward}), which trains a trajectory-conditioned reward model via evidence-grounded SFT and trajectory-level DPO to provide dense per-step process supervision; and
(iv) \emph{Process-Aligned Policy Optimization} (Section~\ref{sec:policy-optimization}), which aggregates action-typed rewards into a balanced trajectory signal and optimizes $\pi_\theta$ through PA-DAPO.

\subsection{Traceable Multi-View Event Bank}
\label{sec:event-memory}
The source units in $\mathcal{H}$ often have incompatible granularities: multi-session dialogues are organized as turns, documents may spread a coherent events across multiple paragraphs, and database records are already atomic. Direct fixed-size chunking can therefore break event boundaries and detach memory entries from their source contexts. 
To preserve event integrity and source traceability, we apply a structure-aware compiler $\mathcal{C}$ to transform the history into a traceable event bank $\mathcal{B}=\mathcal{C}(\mathcal{H})=\{e_m\}_{m=1}^{M}$. Each event is defined as $e_m=(\{x_m^{v}\}_{v\in\mathcal{V}},\ell_m)$, where $\mathcal{V}=\{\mathrm{raw},\mathrm{sum},\mathrm{anc},\mathrm{phr}\}$ denotes four complementary retrieval views and
$\ell_m=(n,i_{\mathrm{start}},i_{\mathrm{end}})$ is a provenance locator that anchors the event to its source unit $h_n$.
Specifically, The compiler first partitions $\mathcal{H}$ along topic-shift boundaries, then constructs four views for each event. The raw view $x_m^{\text{raw}}$ preserves lexical details, summary $x_m^{\text{sum}}$ captures macroscopic semantic coherence, query-anchor $x_m^{\text{anc}}$ encodes hypothetical queries the event can answer, and event-phrase $x_m^{\text{phr}}$ provides a dense, low-noise semantic label.

Since no single representation captures lexical, semantic, and query-intent cues simultaneously, for each search query $u_t$ issued by the $\pi_\theta$, we fuse view-specific rankings via equal-weight Reciprocal Rank Fusion \citep{cormack2009reciprocal}: 
\[
\operatorname{score}(e_m\mid u_t)
=
\sum_{v\in\mathcal{V}}
\frac{1}{\kappa+\operatorname{rank}_{v}(e_m\mid u_t)},
\]
where $\kappa>0$ is a smoothing constant. Dense retrieval is used for $x^{\mathrm{sum}}$ and $x^{\mathrm{phr}}$, BM25 for $x^{\mathrm{anc}}$, and a hybrid of both for $x^{\mathrm{raw}}$. The retrieval operator returns the top-$K_{\mathrm{ret}}$ events as candidate list $\mathcal{R}(u_t)$, with each retaining its locator $\ell_m$ for source-verifiable memory writing and reward evaluation.

\subsection{Evidence-State Transition Modeling}
\label{sec:memory-interaction}
A single retrieval pass is often insufficient when the answer depends on evidence scattered across the history. We therefore formulate memory interaction as a finite-horizon Markov Decision Process (MDP) over evidence states. The memory policy actor $\pi_\theta$ sequentially selects memory operations and the process is specified by a state space $\mathcal{S}$, an action space $\mathcal{A}$, a deterministic transition function $\mathcal{F}$, and a maximum horizon $T_{\max}$.

At step $t\in\{1,\ldots,T\}$, the evidence state is 
$z_t=(q,\mathcal{W}_t,\mathcal{R}_t)\in\mathcal{S}$, where $q$ is the query, $\mathcal{W}_t$ is the current working memory, $\mathcal{R}_t$ is the latest candidate list, and $m_t\in\{\mathrm{decide},\mathrm{extract}\}$ indicates the current decision mode. The memory policy actor samples an action $a_t\sim\pi_\theta(\cdot\mid z_t)$ under a mode-specific action mask, and $c_t=\operatorname{type}(a_t)$ denotes its action type. 
In the \emph{decide} mode, the policy chooses either \textbf{search}$(u_t)$, which issues a search query $u_t$ to the event bank and retrieves $\mathcal{R}(u_t)$, or \textbf{stop}, which terminates the episode.
After a search, the process enters the \emph{extract} mode, where the policy chooses either \textbf{add}$(n_t)$, which appends a source-bound note $n_t$ to the working memory, or \textbf{skip}, which discards the current candidates. This constrained structure prevents invalid operations, such as adding a note before retrieval or issuing a new search before deciding whether to use the current candidates.
The four action types $\{\text{search},\text{add},\text{skip},\text{stop}\}$ induce the following deterministic transitions:
\begin{equation}
z_{t+1}=\mathcal{F}(z_t,a_t;\mathcal{B})=
\begin{cases}
(q,\mathcal{W}_t,\mathcal{R}(u_t)), & c_t=\text{search}\\
(q,\mathcal{W}_t\oplus[n_t],\mathcal{R}_t), & c_t=\text{add}\\
(q,\mathcal{W}_t,()), & c_t=\text{skip}\\
(q,\mathcal{W}_t,\mathcal{R}_t), & c_t=\text{stop}
\end{cases}
\label{eq:state-transition}
\end{equation}
The episode ends when the memory policy actor selects \textbf{stop}, or when the maximum horizon $T_{\max}$ is reached. The final working memory is passed to the frozen generator to produce $\hat{y}=\mathcal{G}(q,\mathcal{W}_T)$. Each interaction trajectory is written as 
$\tau=(z_1,a_1,\ldots,z_T,a_T,z_{T+1})$,
and the collected trajectories form $\mathcal{D}_{\mathrm{traj}}$,
which is used for subsequent reward modeling and policy optimization.

\subsection{Unified Reward Modeling}
\label{sec:reward}
Final-answer correctness provides only terminal feedback and can not identify which memory operations make a trajectory useful. We therefore train a generative reward model $\mathcal{J}_\phi$ to assign dense per-step process rewards in complete memory trajectories. 
Unlike the memory policy actor, $\mathcal{J}_\phi$ evaluates the resulting actions and provides process supervision without affecting state transitions.

\textbf{Stage 1: Evidence-grounded SFT.}
Let $\mathcal{E}^{\star}$ denote the gold evidence spans and $\operatorname{Src}(\cdot)$ the provenance projection that maps an event or note to its referenced raw source spans. 
At step $t$, the uncovered gold evidence is $\mathcal{E}^{\mathrm{miss}}_t=\mathcal{E}^{\star}\setminus\operatorname{Src}(\mathcal{W}_t)$. 
For a search action, the newly exposed gold evidence is  $\Delta_t^{\mathrm{ret}}=\mathcal{E}^{\mathrm{miss}}_t\cap\operatorname{Src}(\mathcal{R}(u_t))$, and for add action, the newly commits gold evidence is $\Delta_t^{\mathrm{write}}=\mathcal{E}^{\mathrm{miss}}_t\cap\operatorname{Src}(n_t)$. 
We use action-specific deterministic rules over these evidence-grounded information to compute a step-level rule score $r_t^{\mathrm{rule}}$ for each action type $c\in\mathcal{A}$. 
For each trajectory $\tau$ produced by the memory policy actor, we serialize the trajectory as $x_\tau=\operatorname{Serialize}(y^{\star},\mathcal{E}^{\star},\tau)$, and fine-tune $\mathcal{J}_\phi$ on the rule reward sequence $\mathbf{v}_b^{\mathrm{rule}}=\{(c_t,r_t^{\mathrm{rule}})\}$:
\begin{equation}
\mathcal{L}_{\mathrm{SFT}}(\phi)=-\frac{1}{\sum_b L_b^{\mathrm{RM}}}\sum_b\sum_t \log p_\phi\!\left(v_{b,t}^{\mathrm{rule}}\mid x_b,\mathbf{v}_{b,<t}^{\mathrm{rule}}\right)
\end{equation}
where $L_b^{\mathrm{RM}}$ the target token length, and $p_\phi$ the autoregressive output distribution of $\mathcal{J}_\phi$.

\textbf{Stage 2: Trajectory-level DPO alignment.}
Rule-derived scores provide action-level supervision based on evidence coverage and provenance, but do not directly assess the semantic quality of an action in its current evidence state. We therefore provide a reasoning-capable model with the full trajectory context $x_\tau$ and have it independently score each step via chain-of-thought reasoning, producing $\mathbf{v}_b^{\mathrm{CoT}}$. We treat $\mathbf{v}_b^+=\mathbf{v}_b^{\mathrm{CoT}}$ as chosen and $\mathbf{v}_b^-=\mathbf{v}_b^{\mathrm{rule}}$ as rejected, forming preference dataset $\mathcal{D}_{\mathrm{pref}}=\{(x_b,\mathbf{v}_b^+,\mathbf{v}_b^-)\}$, and minimize the DPO loss with frozen reference $\phi_{\mathrm{ref}}$ and strength $\beta>0$:
\begin{equation}
\mathcal{L}_{\mathrm{DPO}}(\phi)=-\mathbb{E}\!\left[\log\operatorname{sigmoid}\!\left(\beta\log\frac{p_\phi(\mathbf{v}^{+}\mid x)}{p_{\phi_{\mathrm{ref}}}(\mathbf{v}^{+}\mid x)}-\beta\log\frac{p_\phi(\mathbf{v}^{-}\mid x)}{p_{\phi_{\mathrm{ref}}}(\mathbf{v}^{-}\mid x)}\right)\right]
\end{equation}
After training, $\mathcal{J}_\phi$ directly outputs $\{(c_t,\hat{r}_t)\}_{t=1}^T$, where $\hat{r}_t\in\mathbb{R}$ assesses the semantic quality of action $c_t$ in its current evidence state.

\subsection{Process-Aligned Memory Policy Optimization}
\label{sec:policy-optimization}

AiO-Mem unifies heterogeneous memory action types within a shared evidence-state transition process. We aggregate their step-level rewards into an action-balanced trajectory signal, averaging within action types before integrating across types. For each category $c\in\mathcal{A}$, let $\mathcal{I}_c(\tau)=\{t:c_t=c\}$ be the steps where $c$ occurs, $\overline{r}_c(\tau)=|\mathcal{I}_c(\tau)|^{-1}\!\sum_{t\in\mathcal{I}_c(\tau)}\hat{r}_t$ the within-category mean, and $g_c(r)=(r-r_c^{\mathrm{lo}})/(r_c^{\mathrm{hi}}-r_c^{\mathrm{lo}})$ a normalization to $[0,1]$ using the rubric bounds $r_c^{\mathrm{lo}}$ and $r_c^{\mathrm{hi}}$. Denoting by $\mathcal{A}(\tau)=\{c\in\mathcal{A}:\mathcal{I}_c(\tau)\neq\varnothing\}$ the categories present in $\tau$, the unified trajectory reward is:
\begin{equation}
R_{\mathrm{AiO}}(\tau)=\frac{1}{|\mathcal{A}(\tau)|}\sum_{c\in\mathcal{A}(\tau)}g_c\!\left(\overline{r}_c(\tau)\right)
\end{equation}

\textbf{PA-DAPO.} We optimize the memory policy actor $\pi_\theta$ with our proposed PA-DAPO. Building on DAPO \citep{yu2026dapo}, AiO-Mem extends the loss computation from the conventional $(\texttt{group},\texttt{token})$ structure to a new $(\texttt{group},\texttt{action},\texttt{token})$ hierarchy over trajectories. Formally, let $\mathcal{D}_{\mathrm{RL}}$ denote the distribution of query--history pairs used for policy optimization. For $(q,\mathcal{H})\sim\mathcal{D}_{\mathrm{RL}}$, the old memory policy actor samples $G$ trajectories $\{\tau_j\}$ with rewards $R_j=R_{\mathrm{AiO}}(\tau_j)$. The group-relative advantage $\hat{A}_j=(R_j-\mu)/\sigma$ is computed from the within-group mean $\mu$ and standard deviation $\sigma$. Let $o_{j,k,t}$ be the $t$-th policy-generated token of the $k$-th action step in $\tau_j$, $\mathbf{h}_{j,k,t}$ the preceding context, and $L_{j,k}$ the step's token count. PA-DAPO maximizes:
\begin{equation}
\mathcal{J}_{\mathrm{PA\text{-}DAPO}}(\theta)=\mathbb{E}\!\left[\frac{\sum_{j}\sum_{k}\sum_{t}\Gamma_{j,k,t}(\theta)}{\sum_{j}\sum_{k}L_{j,k}}\right],\quad\sigma>0
\end{equation}
where $\Gamma_{j,k,t}(\theta)=\min(\rho_{j,k,t}\hat{A}_j,\operatorname{clip}(\rho_{j,k,t},1-\epsilon_{\mathrm{low}},1+\epsilon_{\mathrm{high}})\hat{A}_j)$ is the clipped surrogate, $\rho_{j,k,t}(\theta)=\pi_\theta(o_{j,k,t}\mid\mathbf{h}_{j,k,t})/\pi_{\theta_{\mathrm{old}}}(o_{j,k,t}\mid\mathbf{h}_{j,k,t})$ the token-level importance ratio, and $\epsilon_{\mathrm{low}},\epsilon_{\mathrm{high}}$ the clipping bounds. The constraint $\sigma>0$ retains only queries with reward variation, filtering fully converged samples. Operating on an action-balanced trajectory reward rather than binary answer correctness, PA-DAPO provides a graded training signal for end-to-end policy optimization.

\section{Experiment}
\subsection{Experimental Setting}
\paragraph{Datasets and Evaluation}
We evaluate AiO-Mem on three complementary memory benchmarks, including LoCoMo \citep{maharana2024locomo}, MemoryAgentBench \citep{hu2026memoryagentbench}, and PersonaMem \citep{jiang2025personamem}. LoCoMo evaluates long-range conversational memory across multi-session interactions. Following MEM-T~\citep{yue2026memt}, we adopt the same data configuration by splitting the LoCoMo dataset into training, validation, and test sets with a ratio of $1{:}1{:}8$ to ensure a fair comparison.  MemoryAgentBench valuates memory agents under incremental multi-turn interactions across diverse tasks. Following Fine-Mem \citep{ma2026finemem}, we adopt the same evaluation scope and baselines to assess out-of-distribution generalization. PersonaMem further extends evaluation from factual memory to dynamic user profiling. More details are shown in Appendix~\ref{app:datasets-baselines}.

\begingroup
\makeatletter
\providecommand{\ph}[1]{\textcolor{red}{#1}}
\definecolor{headerblue}{RGB}{234,240,250}
\providecolor{srcblue}{RGB}{18,62,138}
\providecolor{estorange}{RGB}{201,110,17}
\providecommand{\src}[1]{\textcolor{srcblue}{#1}}
\providecommand{\est}[1]{\textcolor{estorange}{#1}}
\expandafter\ifx\csname NC@rewrite@C\endcsname\relax
  \newcolumntype{C}[1]{>{\centering\arraybackslash}p{#1}}
\fi
\makeatother

\definecolor{hltop1}{RGB}{140,190,238}
\definecolor{hltop2}{RGB}{214,233,250}
\newcommand{\hlone}[1]{\cellcolor{hltop1}#1}
\newcommand{\hltwo}[1]{\cellcolor{hltop2}#1}

\setlength{\fboxsep}{1pt}

\begin{table}[!t]
\centering
\setlength{\belowcaptionskip}{2pt}
\caption{Main results on \textbf{LoCoMo}.}
\label{tab:dimension_results}
\vspace{-0.3em}
{\fontsize{8pt}{10pt}\selectfont
\setlength{\tabcolsep}{3pt}
\renewcommand{\arraystretch}{0.85}
\begin{tabular}{@{}l *{10}{C{0.99cm}} C{0.99cm}@{}}
\toprule
\multirow{2}{*}{\textbf{Method}}
  & \multicolumn{2}{c}{\textbf{Multi-hop}}
  & \multicolumn{2}{c}{\textbf{Temporal}}
  & \multicolumn{2}{c}{\textbf{Open-domain}}
  & \multicolumn{2}{c}{\textbf{Single-hop}}
  & \multicolumn{2}{c}{\textbf{Overall}} \\
\cmidrule(lr){2-3}\cmidrule(lr){4-5}\cmidrule(lr){6-7}\cmidrule(lr){8-9}\cmidrule(lr){10-11}
  & F1 & B1
  & F1 & B1
  & F1 & B1
  & F1 & B1
  & F1 & B1 \\
\midrule
\multicolumn{11}{@{}l}{\cellcolor{headerblue}\textit{\textbf{Reference}}} \\
Long-Context
  & 23.2 & 16.8 & 19.0 & 13.4 & 13.9 & 10.7 & 40.7 & 31.5 & 31.5 & 23.9 \\
RAG
  & 23.5 & 17.1 & 43.1 & 37.4 & 20.2 & 14.9 & 49.5 & 44.9 & 41.6 & 36.5 \\
\midrule
\multicolumn{11}{@{}l}{\cellcolor{headerblue}\textit{\textbf{Agent Memory Harness}}} \\
MemGPT
  & 16.7 & 14.0 & 12.6 & 10.9 & 11.6 & 9.2 & 14.0 & 11.8 & 14.1 & 11.8 \\
MemoryBank
  & 25.5 & 19.4 & 9.2 & 7.4 & 16.4 & 12.4 & 26.7 & 17.7 & 22.3 & 15.7 \\
Mem0
  & 35.4 & 27.4 & 46.8 & 39.5 & 26.6 & 21.0 & 47.3 & 40.7 & 43.7 & 36.8 \\
MemoryOS
  & 35.2 & 27.3 & 41.0 & 32.7 & 22.1 & 17.9 & 48.4 & 42.6 & 42.8 & 36.3 \\
LightMem
  & 30.8 & 25.8 & 44.7 & 40.7 & 18.9 & 14.4 & 43.8 & 38.8 & 40.0 & 35.3 \\
A-Mem
  & 27.2 & 21.1 & 43.9 & 36.0 & 15.4 & 12.7 & 44.6 & 38.3 & 39.4 & 33.0 \\
GAM
  & 32.2 & 28.7 & 26.3 & 22.5 & 18.5 & 14.5 & 32.2 & 25.5 & 30.2 & 24.8 \\
\midrule
\multicolumn{11}{@{}l}{\cellcolor{headerblue}\textit{\textbf{Trained Method}}} \\
MEM1
  & 19.0 & 15.6 & 30.5 & 23.5 & 14.2 & 11.4 & 27.5 & 22.1 & 25.7 & 20.5 \\
MemAgent
  & 27.9 & 22.7 & 37.9 & 31.9 & 20.3 & 16.5 & 35.9 & 29.6 & 33.8 & 28.0 \\
Mem-T
  & \hltwo{\underline{45.1}} & \hltwo{\underline{36.6}}
  & \hltwo{\underline{65.1}} & \hlone{\textbf{60.1}}
  & \hltwo{\underline{33.0}} & \hltwo{\underline{28.9}}
  & \hltwo{\underline{63.8}} & \hltwo{\underline{58.0}}
  & \hltwo{\underline{58.7}} & \hltwo{\underline{52.6}} \\
\midrule
\multicolumn{11}{@{}l}{\cellcolor{headerblue}\textit{\textbf{Ours}}} \\
AiO-Mem
  & \hlone{\textbf{47.1}} & \hlone{\textbf{38.8}}
  & \hlone{\textbf{66.0}} & \hltwo{\underline{59.4}}
  & \hlone{\textbf{33.7}} & \hlone{\textbf{30.5}}
  & \hlone{\textbf{67.0}} & \hlone{\textbf{61.5}}
  & \hlone{\textbf{61.0}} & \hlone{\textbf{55.0}} \\
\bottomrule
\end{tabular}
}
\end{table}
\vspace{-1.1em}
\endgroup

\paragraph{Baselines.} Twenty systems in three groups. \circled{1} \textit{Reference:} Long-Context and RAG.
\circled{2} \textit{Agent Memory Harness:} MemGPT~\citep{packer2023memgpt}, MemoryBank~\citep{zhong2024memorybank}, Mem0~\citep{chhikara2025mem0}, MemoryOS~\citep{kang2025memoryos}, LightMem~\citep{fang2026lightmem}, A-Mem~\citep{xu2026amem}, GAM~\citep{yan2025gam}, Zep~\citep{rasmussen2025zep}, LangMem~\citep{langchain2025langmem}, EverMemOS~\citep{hu2026evermemos}, and Emotional RAG~\citep{huang2024emotionalrag}. \circled{3} \textit{RL-Trained Method:} MEM1~\citep{zhou2026mem1}, MemAgent~\citep{yu2026memagent}, Mem-T~\citep{yue2026memt}, AtomMem~\citep{huo2026atommem}, Mem-$\alpha$~\citep{wang2025memalpha}, FineMem~\citep{ma2026finemem}, and TrustMem~\citep{yang2026trustmem}. Details are provided in Appendix~\ref{app:datasets-baselines}.

\paragraph{Experimental Details.}
We instantiate AiO-Mem with Qwen3-8B as the reward model and Qwen3-4B as the memory policy actor. BGE-M3\footnote{https://huggingface.co/BAAI/bge-m3.} and BM25 are used for multi-view event retrieval. For policy training, we set $K_{\mathrm{ret}}=10$, $T_{\max}=8$. For reward-model SFT, we train for 2 epochs with a batch size of 8 and a learning rate of $1\times10^{-4}$. Trajectory-level DPO is then performed for 1 epoch with a batch size of 8, a learning rate of $2\times10^{-6}$, and $\beta=0.1$. For PA-DAPO, we sample 8 trajectories per query and optimize the memory policy with a learning rate of $2\times10^{-5}$. More training setup and parameter configurations are listed in Appendix~\ref{app:implementation-details}.

\begin{figure}[t]
    \centering
    \includegraphics[width=\linewidth]{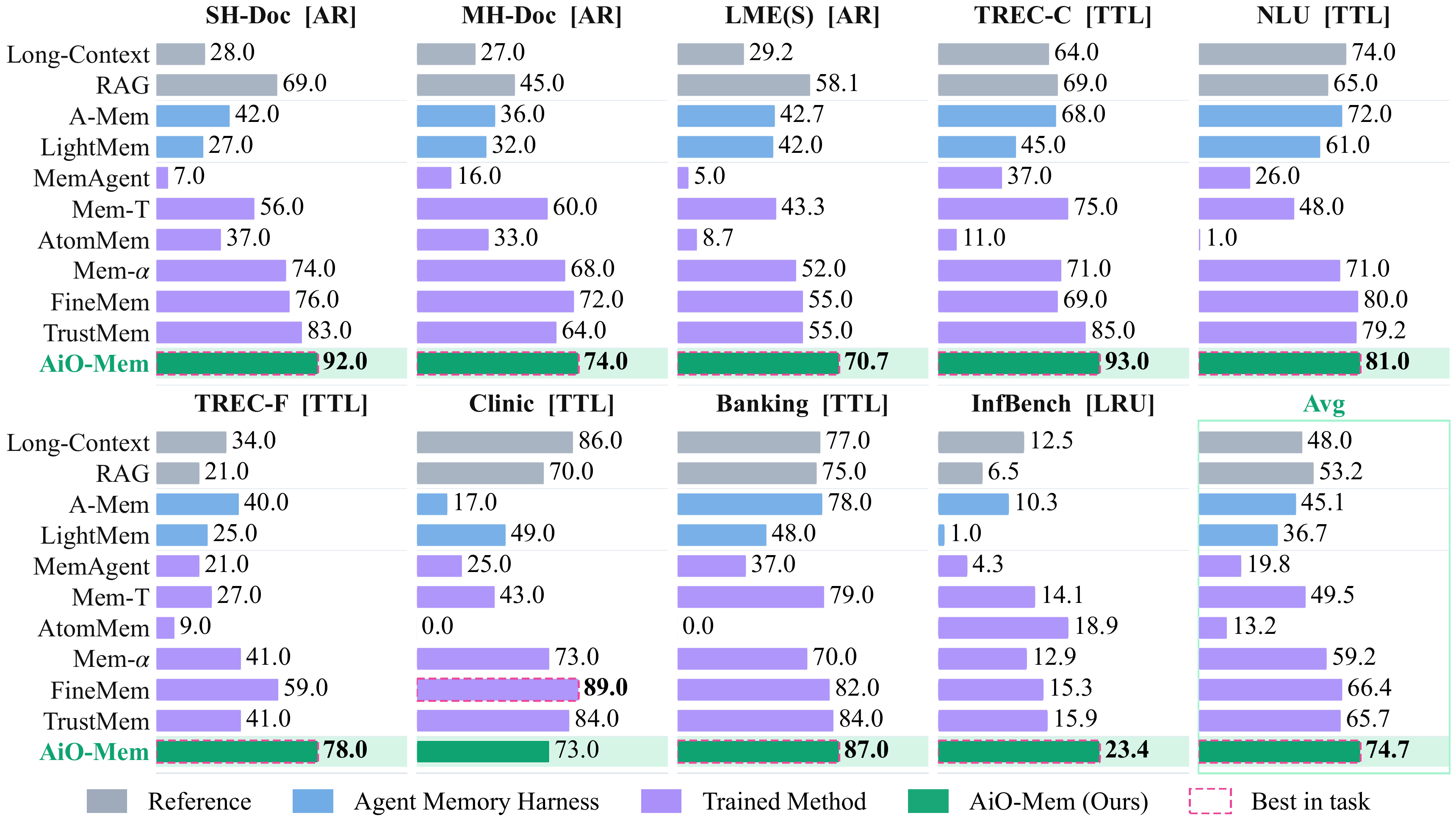}
    \caption{Main results on \textbf{MemoryAgentBench}.}
    \label{fig:main_results_memoryagentbench}
    \vspace{-9pt}
\end{figure}

\subsection{Main Results}

\paragraph{LoCoMo.}
As shown in Table~\ref{tab:dimension_results}, , AiO-Mem achieves the strongest overall performance and consistently outperforms competing methods across diverse question types. The gains span both evidence-intensive reasoning and direct memory queries, reflecting a coherent improvement pattern rather than isolated advantages on specific categories. In contrast, existing memory systems exhibit more uneven performance across different reasoning demands. These results suggest that \textbf{jointly modeling heterogeneous memory actions under shared evidence states and unified process supervision enables more reliable memory interaction throughout the reasoning process}.

\paragraph{MemoryAgentBench.}
As shown in Figure~\ref{fig:main_results_memoryagentbench}, AiO-Mem achieves the strongest overall performance across MemoryAgentBench, with gains that form a consistent pattern across retrieval, test-time learning, and long-range understanding. It achieves an average score of 74.7 across the nine tasks, outperforming Fine-Mem, the strongest baseline, by 8.3 points. Compared with the best competing result on each task, AiO-Mem improves LME(S), TREC-F, and InfBench by 12.6, 19.0, and 4.5 points, respectively. Such gains across heterogeneous task formats demonstrate \textbf{strong out-of-distribution generalization beyond the LoCoMo training setting}. This further supports the design of AiO-Mem for generalizable memory interaction.

\paragraph{PersonaMem.} AiO-Mem achieves the best overall performance on PersonaMem, surpassing the strongest baseline by 3.9 points and \textbf{demonstrating stronger long-term user profiling and personalization}. More details are shown in Appendix~\ref{app:datasets-baselines}.

\subsection{Ablation Study}

\begingroup
\makeatletter
\providecommand{\ph}[1]{\textcolor{red}{#1}}
\providecommand{\drop}[1]{%
  \textcolor{red}{\makebox[2.7em][r]{\tiny(#1)}}%
}
\definecolor{headerblue}{RGB}{234,240,250}
\providecolor{srcblue}{RGB}{18,62,138}
\providecolor{estorange}{RGB}{201,110,17}
\providecommand{\src}[1]{\textcolor{srcblue}{#1}}
\providecommand{\est}[1]{\textcolor{estorange}{#1}}

\expandafter\ifx\csname NC@rewrite@C\endcsname\relax
  \newcolumntype{C}[1]{>{\centering\arraybackslash}p{#1}}
\fi
\makeatother

\setlength{\fboxsep}{1pt}

\begin{table*}[!t]
\centering
\setlength{\belowcaptionskip}{2pt}
\caption{%
Ablation on \textbf{LoCoMo}.
Row abbreviations:
Event Comp.~=~Event Compilation,
Multi-view Ret.~=~Multi-view Retrieval,
WM Constr.~=~Working-Memory Construction,
Evid. SFT~=~Evidence-Grounded SFT,
and Traj. DPO~=~Trajectory-Level DPO.
\textcolor{red}{Red} values denote drops over AiO-Mem.%
}
\label{tab:ablation_results}
\vspace{-0.3em}

{\fontsize{8pt}{10pt}\selectfont
\setlength{\tabcolsep}{2.5pt}
\renewcommand{\arraystretch}{0.85}

\resizebox{\linewidth}{!}{%
\begin{tabular}{@{}l *{8}{C{0.92cm}} *{2}{C{1.45cm}}@{}}
\toprule
\multirow{2}{*}{\textbf{Method}}
& \multicolumn{2}{c}{\textbf{Multi-hop}}
& \multicolumn{2}{c}{\textbf{Temporal}}
& \multicolumn{2}{c}{\textbf{Open-domain}}
& \multicolumn{2}{c}{\textbf{Single-hop}}
& \multicolumn{2}{c}{\textbf{Overall}} \\
\cmidrule(lr){2-3}
\cmidrule(lr){4-5}
\cmidrule(lr){6-7}
\cmidrule(lr){8-9}
\cmidrule(lr){10-11}
& F1 & B1 & F1 & B1 & F1 & B1 & F1 & B1 & F1 & B1 \\
\midrule

\textit{Base Model}
& 39.3 & 30.9 & 65.0 & 58.4 & 25.9 & 21.9 & 61.3 & 56.1
& 55.8\drop{-5.2} & 49.8\drop{-5.2} \\

\midrule
\multicolumn{11}{@{}l}{\textit{\textbf{Memory Representation and Interaction Ablations}}} \\

\textit{w/o Event Comp.}
& 43.2 & 34.6 & 66.9 & 61.0 & 27.2 & 22.6 & 64.3 & 58.7
& 58.6\drop{-2.4} & 52.5\drop{-2.5} \\

\textit{w/o Multi-view Ret.}
& 41.3 & 33.4 & 65.8 & 60.4 & 29.5 & 24.4 & 64.2 & 58.6
& 58.1\drop{-2.9} & 52.1\drop{-2.9} \\

\textit{w/o WM Constr.}
& 33.2 & 26.4 & 57.0 & 50.9 & 24.3 & 20.0 & 57.6 & 52.3
& 50.9\drop{-10.1} & 45.2\drop{-9.8} \\

\midrule
\multicolumn{11}{@{}l}{\textit{\textbf{Unified Reward Modeling Ablations}}} \\

\textit{w/o Evid. SFT}
& 41.4 & 33.6 & 68.4 & 60.6 & 28.6 & 24.5 & 62.9 & 57.4
& 57.9\drop{-3.1} & 51.6\drop{-3.4} \\

\textit{w/o Traj. DPO}
& 43.4 & 35.2 & 66.5 & 59.5 & 34.2 & 29.9 & 64.0 & 58.4
& 58.8\drop{-2.2} & 52.6\drop{-2.4} \\

\midrule
\rowcolor{headerblue}
AiO-Mem
& \textbf{47.1} & \textbf{38.8}
& \textbf{66.0} & \textbf{59.4}
& \textbf{33.7} & \textbf{30.5}
& \textbf{67.0} & \textbf{61.5}
& \textbf{61.0} & \textbf{55.0} \\

\bottomrule
\end{tabular}%
}
}
\end{table*}
\endgroup
To examine the contribution of each component in AiO-Mem, we conduct ablation experiments by removing individual modules, as shown in Table~\ref{tab:ablation_results}. Removing Event Compilation (\textit{w/o Event Comp.}) or Multi-view Retrieval (\textit{w/o Multi-view Ret.}) consistently degrades performance, showing that structured event organization and complementary retrieval cues provide a reliable evidence basis for memory interaction. Removing Working-Memory Construction (\textit{w/o WM Constr.}) causes the most substantial degradation, highlighting its role in progressively transforming retrieved evidence into reasoning-sufficient working memory. For Unified Reward Modeling, removing Evidence-Grounded SFT (\textit{w/o Evid. SFT}) weakens performance, demonstrating the importance of evidence-based supervision for establishing reliable process rewards. Further removing Trajectory-Level DPO (\textit{w/o Traj. DPO}) also leads to consistent degradation, showing that trajectory-level alignment improves reward estimation beyond rule-based supervision. Overall, AiO-Mem achieves the best performance by jointly integrating these components within a unified memory-interaction framework.

\begin{figure}[t]
    \centering
    \includegraphics[width=\linewidth]{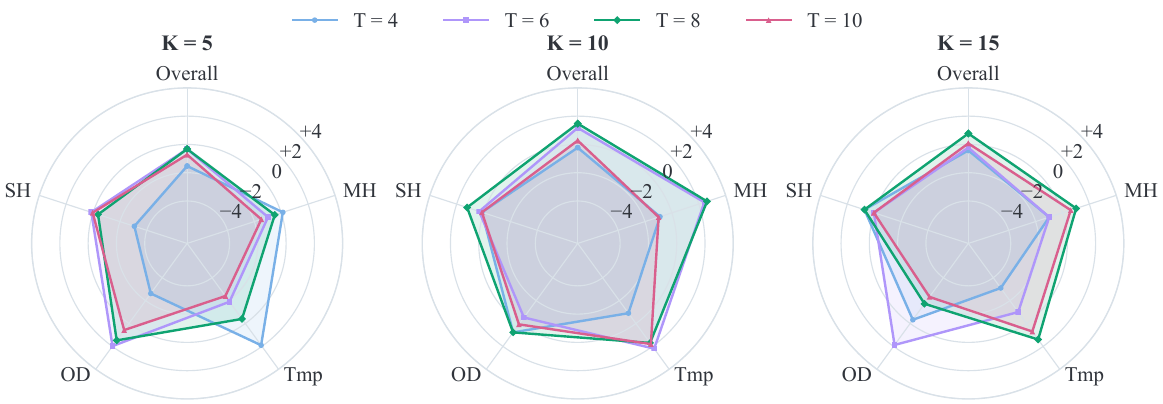}
    \caption{Parameter analysis on \textbf{LoCoMo} over top-$K$ ($K\in\{5,10,15\}$) and maximum interaction steps $T_{\max}\in\{4,6,8,10\}$, reporting F1 across four question types (MH~=~Multi-hop, Tmp~=~Temporal, OD~=~Open-domain, SH~=~Single-hop) and overall. To facilitate comparison, all values are normalized for visualization on the chart.}
    \label{fig:param_composite}
    \vspace{-1.1em}
\end{figure}

\subsection{Analysis of Complexity and Sensitivity}
\paragraph{Complexity Analysis.}
Existing memory systems repeatedly summarize and rewrite accumulated memories, incurring a history-dependent cost of $O(N\ell+NM_1R_1)\mathcal{C}_{\mathrm{LLM}}$. In contrast, AiO-Mem decouples persistent storage from query-time working-memory construction. The event bank $\mathcal{B}$ is built in a single pass, while query-time reasoning is bounded by $K_{\mathrm{ret}}$ and $T_{\max}$, yielding $O(N\ell+T_{\max}K_{\mathrm{ret}}\ell_e)\mathcal{C}_{\mathrm{LLM}}$. By bounding the working context to a constant size, AiO-Mem transforms a history-dependent scaling cost into a local, horizon-bounded overhead that remains invariant to the stream length. Empirically, this decoupling mechanism reduces retrieval and construction token consumption by 79.9\% and 73.8\% respectively, bypassing the active memory maintenance bottleneck of traditional architectures. Detailed derivations are provided in Appendix~\ref{app:complexity}.

\paragraph{Sensitivity Analysis.}
We further examine the robustness of AiO-Mem to two key hyperparameters, the retrieval width $K_{\mathrm{ret}}$ and maximum interaction horizon $T_{\max}$. As shown in Figure~\ref{fig:param_composite}, performance remains relatively stable across  configurations, while increasing either parameter yields no consistent gains. A moderate retrieval width provides sufficient evidence coverage without introducing excessive irrelevant candidates, while an appropriate interaction horizon enables effective multi-step memory interaction without redundant actions. The configuration $(K_{\mathrm{ret}},T_{\max})=(10,8)$ achieves the strongest overall performance, suggesting that AiO-Mem benefits from a balanced retrieval and interaction budget. Detailed F1 and B1 results are provided in Appendix~\ref{app:sensitivity}.

\begin{figure}[t]
    \centering
    \includegraphics[width=\linewidth]{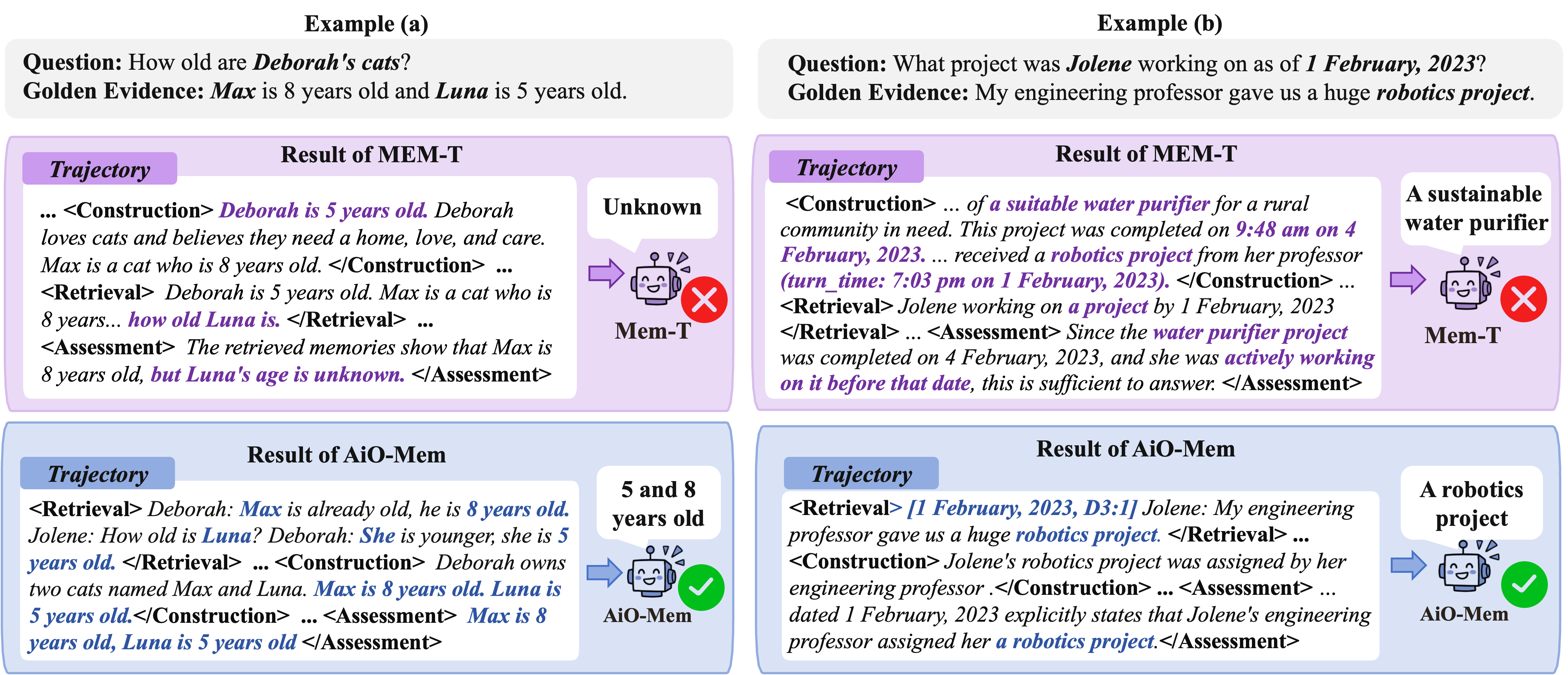}
    \caption{Two cases between the MEM-T and our AiO-Mem method.}
    \label{fig:case}
    \vspace{-1.5em}
\end{figure}

\subsection{Case Study}
To further illustrate the advantage of AiO-Mem over Mem-T, Figure~\ref{fig:case} presents two memory-interaction cases. In Example~(a), Mem-T incorrectly binds ``she is 5 years old'' to Deborah rather than Luna during memory construction. Although relevant evidence is retrieved, the incorrect binding is retained in memory and leads to an incorrect assessment. In contrast, AiO-Mem retrieves source-grounded evidence and constructs a working-memory note that preserves the correct entity relation. In Example~(b), Mem-T conflates the temporally distinct robotics and water-purifier projects, causing the assessment to overlook the query-time cutoff and select evidence from a later event. AiO-Mem instead retrieves dated evidence and preserves its temporal relation during working-memory construction, leading to the correct answer. These cases show that AiO-Mem improves memory interaction by jointly coordinating retrieval and construction under shared evidence states and process-level supervision, preventing errors from propagating into later reasoning.

\section{Additional Ablations in Appendix}
We further examine AiO-Mem across the following key dimensions. \textbf{(i) Multi-View Retrieval Complementarity and Sensitivity.} The four retrieval views provide complementary evidence coverage, jointly covering 97.03\% of question--evidence pairs while remaining robust to retrieval configurations and context cost (\textbf{\underline{Appendix~\ref{app:multiview-complementarity}}}, \textbf{\underline{Appendix~\ref{app:inference-retrieval-sensitivity}}}).\textbf{(ii) Unified Reward Alignment.} The trained reward model shows stronger correlations with evidence Recall and Precision, better reflecting evidence-grounded trajectory quality (\textbf{\underline{Appendix~\ref{app:reward-alignment}}}). \textbf{(iii) Trajectory-level DPO Alignment Study.} We visualize how trajectory-level DPO distinguishes trajectories with identical Evidence-ID metrics through finer-grained semantic judgments(\textbf{\underline{Appendix~\ref{app:reward-case-study}}}). \textbf{(iv) Training Dynamics.} Overall, consistent gains across Search, Extract, and Answer-Decision rewards indicate stable PA-DAPO training and coordinated step-wise optimization(\textbf{\underline{Appendix~\ref{app:reward-dynamics}}}). \textbf{(v) Process-Aligned Evidence Flow.} PA-DAPO improves evidence retrieval, working-memory quality, and Answer F1 across the full pipeline (\textbf{\underline{Appendix~\ref{app:evidence-flow}}}).

\section{Conclusion}
Existing memory systems often optimize retrieval and memory construction separately, making it difficult to coordinate heterogeneous memory operations under a shared objective. To address this, we introduce AiO-Mem, which unifies retrieval, writing, filtering, and termination as transitions over a shared evidence state, and learns a trajectory-conditioned reward model to provide process-level supervision for memory-policy optimization. Experiments across long-context memory benchmarks show that AiO-Mem consistently outperforms both static memory systems and RL-trained memory agents, with strong out-of-distribution generalization beyond conversational QA. These results highlight the promise of an All-in-One formulation for learning reliable and generalizable memory interaction in LLM agents.


\clearpage
\subsection*{AI use statement}

LLMs were used to generate data and to evaluate memory-interaction trajectories as judges, as described in the paper. The core method development in this paper does not involve LLM usage.  We take responsibility for the final content of this work, including text, claims or artifacts produced with the aid of generative AI.

\subsection*{Ethics statement}

AiO-Mem organizes long interaction histories into retrievable events and
constructs query-specific working memory. Although this can improve an
agent's ability to use historical evidence, such histories may contain
sensitive information. Retaining source-linked events may increase the risk
of inappropriate access to or disclosure of that information. Inaccurate
or biased information in a history may also be retrieved and propagated
into later answers. Our experiments use the existing LoCoMo,
MemoryAgentBench, and PersonaMem benchmarks. Deployment with real user
histories would require appropriate safeguards for user consent, data
access, retention and deletion, as well as mechanisms for identifying and
correcting inaccurate memories.

\subsection*{Reproducibility statement}

The method is described in Section~\ref{sec:method}. Appendix~\ref{app:experimental-details} provides the experimental details. Prompts and memory-agent tool calls are provided
in Appendix~\ref{app:prompts}. To support reproducibility, we will release our code on GitHub .

\bibliography{iclr2027_conference}

@article{openai2023gpt4,
  title   = {{GPT-4} Technical Report},
  author  = {OpenAI},
  journal = {arXiv preprint arXiv:2303.08774},
  year    = {2023}
}

@article{touvron2023llama,
  title   = {{LLaMA}: Open and Efficient Foundation Language Models},
  author  = {Touvron, Hugo and Lavril, Thibaut and Izacard, Gautier and
             Martinet, Xavier and Lachaux, Marie-Anne and Lacroix, Timoth{\'e}e and
             Rozi{\`e}re, Baptiste and Goyal, Naman and Hambro, Eric and
             Azhar, Faisal and Rodriguez, Aurelien and Joulin, Armand and
             Grave, Edouard and Lample, Guillaume},
  journal = {arXiv preprint arXiv:2302.13971},
  year    = {2023}
}

@article{liu2024lost,
  title   = {Lost in the Middle: How Language Models Use Long Contexts},
  author  = {Liu, Nelson F. and Lin, Kevin and Hewitt, John and
             Paranjape, Ashwin and Bevilacqua, Michele and Petroni, Fabio and
             Liang, Percy},
  journal = {Transactions of the Association for Computational Linguistics (TACL)},
  volume  = {12},
  pages   = {157--173},
  year    = {2024}
}

@article{lewis2020rag,
  title={Retrieval-augmented generation for knowledge-intensive nlp tasks},
  author={Lewis, Patrick and Perez, Ethan and Piktus, Aleksandra and Petroni, Fabio and Karpukhin, Vladimir and Goyal, Naman and K{\"u}ttler, Heinrich and Lewis, Mike and Yih, Wen-tau and Rockt{\"a}schel, Tim and others},
  journal={Advances in neural information processing systems},
  volume={33},
  pages={9459--9474},
  year={2020}
}

@inproceedings{guu2020realm,
  title={Retrieval augmented language model pre-training},
  author={Guu, Kelvin and Lee, Kenton and Tung, Zora and Pasupat, Panupong and Chang, Mingwei},
  booktitle={International conference on machine learning},
  pages={3929--3938},
  year={2020},
  organization={PMLR}
}

@article{edge2024graphrag,
  title={From local to global: A graph rag approach to query-focused summarization},
  author={Edge, Darren and Trinh, Ha and Cheng, Newman and Bradley, Joshua and Chao, Alex and Mody, Apurva and Truitt, Steven and Metropolitansky, Dasha and Ness, Robert Osazuwa and Larson, Jonathan},
  journal={arXiv preprint arXiv:2404.16130},
  year={2024}
}

@article{gutierrez2024hipporag,
  title={Hipporag: Neurobiologically inspired long-term memory for large language models},
  author={Guti{\'e}rrez, Bernal J and Shu, Yiheng and Gu, Yu and Yasunaga, Michihiro and Su, Yu},
  journal={Advances in neural information processing systems},
  volume={37},
  pages={59532--59569},
  year={2024}
}

@article{sanmartin2024kgrag,
  title={Kg-rag: Bridging the gap between knowledge and creativity},
  author={Sanmartin, Diego},
  journal={arXiv preprint arXiv:2405.12035},
  year={2024}
}

@inproceedings{huang2024emotionalrag,
  title={Emotional RAG: Enhancing role-playing agents through emotional retrieval},
  author={Huang, Le and Lan, Hengzhi and Sun, Zijun and Shi, Chuan and Bai, Ting},
  booktitle={2024 IEEE International Conference on Knowledge Graph (ICKG)},
  pages={120--127},
  year={2024},
  organization={IEEE}
}

@article{packer2023memgpt,
  title={Memgpt: Towards llms as operating systems},
  author={Packer, Charles and Wooders, Sarah and Lin, Kevin and Fang, Vivian and Patil, Shishir G and Stoica, Ion and Gonzalez, Joseph E},
  journal={arXiv preprint arXiv:2310.08560},
  year={2023}
}

@inproceedings{zhong2024memorybank,
  title={Memorybank: Enhancing large language models with long-term memory},
  author={Zhong, Wanjun and Guo, Lianghong and Gao, Qiqi and Ye, He and Wang, Yanlin},
  booktitle={Proceedings of the AAAI conference on artificial intelligence},
  volume={38},
  number={17},
  pages={19724--19731},
  year={2024}
}

@article{chhikara2025mem0,
  title={Mem0: Building production-ready ai agents with scalable long-term memory},
  author={Chhikara, Prateek and Khant, Dev and Aryan, Saket and Singh, Taranjeet and Yadav, Deshraj},
  journal={arXiv preprint arXiv:2504.19413},
  year={2025}
}

@inproceedings{kang2025memoryos,
  title={Memory os of ai agent},
  author={Kang, Jiazheng and Ji, Mingming and Zhao, Zhe and Bai, Ting},
  booktitle={Proceedings of the 2025 Conference on Empirical Methods in Natural Language Processing},
  pages={25972--25981},
  year={2025}
}

@inproceedings{fang2026lightmem,
  title={Lightmem: Lightweight and efficient memory-augmented generation},
  author={Fang, Jizhan and Deng, Xinle and Xu, Haoming and Jiang, Ziyan and Tang, Yuqi and Xu, Ziwen and Deng, Shumin and Yao, Yunzhi and Wang, Mengru and Qiao, Shuofei and others},
  booktitle={International Conference on Learning Representations},
  volume={2026},
  pages={98706--98729},
  year={2026}
}

@article{xu2026amem,
  title={A-mem: Agentic memory for llm agents},
  author={Xu, Wujiang and Liang, Zujie and Mei, Kai and Gao, Hang and Tan, Juntao and Zhang, Yongfeng},
  journal={Advances in Neural Information Processing Systems},
  volume={38},
  pages={17577--17604},
  year={2026}
}

@article{rasmussen2025zep,
  title={Zep: a temporal knowledge graph architecture for agent memory},
  author={Rasmussen, Preston and Paliychuk, Pavlo and Beauvais, Travis and Ryan, Jack and Chalef, Daniel},
  journal={arXiv preprint arXiv:2501.13956},
  year={2025}
}

@inproceedings{hu2026evermemos,
  title={Evermemos: A self-organizing memory operating system for structured long-horizon reasoning},
  author={Hu, Chuanrui and Gao, Xingze and Zhou, Zuyi and Xu, Dannong and Bai, Yi and Li, Xintong and Zhang, Hui and Li, Tong and Zhang, Chong and Bing, Lidong and others},
  booktitle={Proceedings of the 64th Annual Meeting of the Association for Computational Linguistics (Volume 1: Long Papers)},
  pages={45836--45853},
  year={2026}
}

@misc{langchain2025langmem,
  author       = {{LangChain}},
  title        = {{LangMem}: Long-Term Memory for Language Agents},
  year         = {2025},
  howpublished = {\url{https://github.com/langchain-ai/langmem}},
  note         = {Official software repository and documentation, accessed 2026-09-19}
}

@article{yan2025gam,
  title={General agentic memory via deep research},
  author={Yan, BY and Li, Chaofan and Qian, Hongjin and Lu, Shuqi and Liu, Zheng},
  journal={arXiv preprint arXiv:2511.18423},
  year={2025}
}

@article{jin2025searchr1,
  title={Search-r1: Training llms to reason and leverage search engines with reinforcement learning},
  author={Jin, Bowen and Zeng, Hansi and Yue, Zhenrui and Yoon, Jinsung and Arik, Sercan and Wang, Dong and Zamani, Hamed and Han, Jiawei},
  journal={arXiv preprint arXiv:2503.09516},
  year={2025}
}

@article{song2025r1searcher,
  title={R1-searcher: Incentivizing the search capability in llms via reinforcement learning},
  author={Song, Huatong and Jiang, Jinhao and Min, Yingqian and Chen, Jie and Chen, Zhipeng and Zhao, Wayne Xin and Fang, Lei and Wen, Ji-Rong},
  journal={arXiv preprint arXiv:2503.05592},
  year={2025}
}

@article{jiang2025deepretrieval,
  title={Deepretrieval: Hacking real search engines and retrievers with large language models via reinforcement learning},
  author={Jiang, Pengcheng and Lin, Jiacheng and Cao, Lang and Tian, Runchu and Kang, SeongKu and Wang, Zifeng and Sun, Jimeng and Han, Jiawei},
  journal={arXiv preprint arXiv:2503.00223},
  year={2025}
}

@inproceedings{zhou2026mem1,
  title={Mem1: Learning to synergize memory and reasoning for efficient long-horizon agents},
  author={Zhou, Zijian and Qu, Ao and Wu, Zhaoxuan and Kim, Sunghwan and Prakash, Alok and Rus, Daniela and Low, Bryan Kian Hsiang and Liang, Paul},
  booktitle={International Conference on Learning Representations},
  volume={2026},
  pages={58413--58438},
  year={2026}
}

@inproceedings{yu2026memagent,
  title={Memagent: Reshaping long-context llm with multi-conv rl-based memory agent},
  author={Yu, Hongli and Chen, Tinghong and Feng, Jiangtao and Chen, Jiangjie and Dai, Weinan and Yu, Qiying and Zhang, Ya-Qin and Ma, Wei-Ying and Liu, Jingjing and Wang, Mingxuan and others},
  booktitle={International Conference on Learning Representations},
  volume={2026},
  pages={39458--39486},
  year={2026}
}

@inproceedings{yan2026memoryr1,
  title={Memory-r1: Enhancing large language model agents to manage and utilize memories via reinforcement learning},
  author={Yan, Sikuan and Yang, Xiufeng and Huang, Zuchao and Nie, Ercong and Ding, Zifeng and Li, Zonggen and Ma, Xiaowen and Bi, Jinhe and Kersting, Kristian and Pan, Jeff Z and others},
  booktitle={Proceedings of the 64th Annual Meeting of the Association for Computational Linguistics (Volume 1: Long Papers)},
  pages={12805--12825},
  year={2026}
}

@article{wang2025memalpha,
  title={Mem-$\{$$\backslash$alpha$\}$: Learning memory construction via reinforcement learning},
  author={Wang, Yu and Takanobu, Ryuichi and Liang, Zhiqi and Mao, Yuzhen and Hu, Yuanzhe and McAuley, Julian and Wu, Xiaojian},
  journal={arXiv preprint arXiv:2509.25911},
  year={2025}
}

@misc{huo2026atommem,
      title={AtomMem : Learnable Dynamic Agentic Memory with Atomic Memory Operation}, 
      author={Yupeng Huo and Yaxi Lu and Zhong Zhang and Haotian Chen and Yankai Lin},
      year={2026},
      eprint={2601.08323},
      archivePrefix={arXiv},
      primaryClass={cs.AI},
      url={https://arxiv.org/abs/2601.08323}, 
}

@inproceedings{ma2026finemem,
  title={Fine-Mem: Fine-Grained Feedback Alignment for Long-Horizon Memory Management},
  author={Ma, Weitao and Feng, Xiaocheng and Huang, Lei and Feng, Xiachong and Ma, Zhanyu and Xu, Jun and Gao, Jiuchong and Hao, Jinghua and He, Renqing and Qin, Bing},
  booktitle={Proceedings of the 64th Annual Meeting of the Association for Computational Linguistics (Volume 1: Long Papers)},
  pages={19666--19684},
  year={2026}
}

@article{yang2026trustmem,
  title={TRUSTMEM: Learning Trustworthy Memory Consolidation for LLM Agents with Long-Term Memory},
  author={Yang, Tianyu and Paul, Sudipta and Srinivasan, Vijay and Kulkarni, Vivek and Chappidi, Srinivas},
  journal={arXiv preprint arXiv:2606.25161},
  year={2026}
}

@article{yue2026memt,
  title={Mem-t: Densifying rewards for long-horizon memory agents},
  author={Yue, Yanwei and Peng, Boci and Fan, Xuanbo and Guo, Jiaxin and Li, Qiankun and Zhang, Yan},
  journal={arXiv preprint arXiv:2601.23014},
  year={2026}
}

@inproceedings{maharana2024locomo,
  title={Evaluating very long-term conversational memory of llm agents},
  author={Maharana, Adyasha and Lee, Dong-Ho and Tulyakov, Sergey and Bansal, Mohit and Barbieri, Francesco and Fang, Yuwei},
  booktitle={Proceedings of the 62nd Annual Meeting of the Association for Computational Linguistics (Volume 1: Long Papers)},
  pages={13851--13870},
  year={2024}
}

@inproceedings{hu2026memoryagentbench,
  title={Evaluating memory in llm agents via incremental multi-turn interactions},
  author={Hu, Yuanzhe and Wang, Yu and McAuley, Julian},
  booktitle={International Conference on Learning Representations},
  volume={2026},
  pages={156259--156291},
  year={2026}
}

@article{jiang2025personamem,
  title={Know me, respond to me: Benchmarking llms for dynamic user profiling and personalized responses at scale},
  author={Jiang, Bowen and Hao, Zhuoqun and Cho, Young-Min and Li, Bryan and Yuan, Yuan and Chen, Sihao and Ungar, Lyle and Taylor, Camillo J and Roth, Dan},
  journal={arXiv preprint arXiv:2504.14225},
  year={2025}
}

@misc{calflops,
  author       = {Xiaoju Ye},
  title        = {calflops: a FLOPs and Params calculate tool for neural networks in pytorch framework},
  year         = {2023},
  howpublished = {\url{https://github.com/MrYxJ/calculate-flops.pytorch}}
}

@article{johnson1967hierarchical,
  title={Hierarchical clustering schemes},
  author={Johnson, Stephen C},
  journal={Psychometrika},
  volume={32},
  number={3},
  pages={241--254},
  year={1967},
  publisher={Springer-Verlag}
}

@article{ward1963hierarchical,
  title   = {Hierarchical Grouping to Optimize an Objective Function},
  author  = {Ward, Joe H.},
  journal = {Journal of the American Statistical Association},
  volume  = {58},
  number  = {301},
  pages   = {236--244},
  year    = {1963}
}

@inproceedings{cormack2009reciprocal,
  title={Reciprocal rank fusion outperforms condorcet and individual rank learning methods},
  author={Cormack, Gordon V and Clarke, Charles LA and Buettcher, Stefan},
  booktitle={Proceedings of the 32nd international ACM SIGIR conference on Research and development in information retrieval},
  pages={758--759},
  year={2009}
}

@article{yu2026dapo,
  title={Dapo: An open-source llm reinforcement learning system at scale},
  author={Yu, Qiying and Zhang, Zheng and Zhu, Ruofei and Yuan, Yufeng and Zuo, Xiaochen and Yue, Yu and Dai, Weinan and Fan, Tiantian and Liu, Gaohong and Liu, Lingjun and others},
  journal={Advances in Neural Information Processing Systems},
  volume={38},
  pages={113222--113244},
  year={2026}
}
\bibliographystyle{iclr2027_conference}

\newpage

\appendix
\section{Appendix Overview and Organization}
This appendix provides supplementary details to support and extend the main paper. The organization of the appendix is as follows:

\begin{enumerate}
    \item \textbf{Additional Related Work (Section~\ref{app:related_work}):}
    This section provides further discussion of memory-augmented large language models and reinforcement learning methods for memory policies, with particular emphasis on persistent memory management and process-level memory optimization.

    \item \textbf{Experimental Setup and Supplementary Experiments 
    (Sections~\ref{app:experimental-details} and~\ref{app:additional-experiments}):}
    These sections describe the datasets, baselines, and implementation details, and provide additional experiments on parameter sensitivity, PersonaMem, multi-view retrieval complementarity, and inference-time retrieval configurations.

    \item \textbf{Reward, Optimization, and Efficiency Analysis 
    (Sections~\ref{app:reward-optimization} and~\ref{app:complexity}):}
    These sections further analyze reward alignment, qualitative reward cases, PA-DAPO training dynamics, evidence flow through the memory pipeline, as well as the computational complexity and inference efficiency of AiO-Mem.

    \item \textbf{Prompts and Tool Calls (Section~\ref{app:prompts}):}
    This section provides the memory-agent tool definitions and prompt templates used for memory interaction and stepwise reward modeling, supporting reproducibility of the proposed framework.
\end{enumerate}

\section{Related Work}
\label{app:related_work}
\subsection{Additional Discussion on Memory-Augmented Large Language Models}

Memory-augmented language models have evolved from external retrieval toward persistent and structured memory management. Early retrieval-augmented methods, including RAG~\citep{lewis2020rag} and REALM~\citep{guu2020realm}, introduce non
-parametric knowledge through dense retrieval, while subsequent systems such as GraphRAG~\citep{edge2024graphrag}, HippoRAG~\citep{gutierrez2024hipporag}, KG-RAG~\citep{sanmartin2024kgrag}, and Emotional RAG~\citep{huang2024emotionalrag} enrich retrieval with graph structures, multi-step exploration, or task-specific similarity signals. These methods primarily focus on improving how external evidence is accessed for a given query. Another line of work develops persistent memory across interactions. MemGPT~\citep{packer2023memgpt}, MemoryBank~\citep{zhong2024memorybank}, and Mem0~\citep{chhikara2025mem0} introduce explicit mechanisms for storing, updating, and maintaining long-term information, while MemoryOS~\citep{kang2025memoryos} and LightMem~\citep{fang2026lightmem} improve scalability through hierarchical storage and lightweight consolidation. More recent systems, including A-Mem~\citep{xu2026amem}, Zep~\citep{rasmussen2025zep}, EverMemOS~\citep{hu2026evermemos}, LangMem~\citep{langchain2025langmem}, and GAM~\citep{yan2025gam}, further explore atomic memory units, temporal structures, background consolidation, and iterative evidence gathering. These methods substantially expand the capabilities of long-term memory systems in construction, organization, updating, and retrieval. However, the individual operations are still largely coordinated through predefined workflows, prompts, or heuristic rules, rather than being \textbf{jointly optimized according to their contribution to downstream evidence acquisition}.

\subsection{Additional Discussion on Reinforcement Learning for Memory Policies}

Reinforcement learning has increasingly been introduced to replace manually specified memory behaviors with learned policies. Early efforts mainly focus on retrieval, where Search-R1~\citep{jin2025searchr1}, R1-Searcher~\citep{song2025r1searcher}, and DeepRetrieval~\citep{jiang2025deepretrieval} optimize multi-step search, tool invocation, or query generation through task- or retrieval-oriented rewards. Another line applies RL to recurrent memory states: MEM1~\citep{zhou2026mem1} and MemAgent~\citep{yu2026memagent} learn to continually compress or overwrite bounded memory representations over long contexts and propagate downstream rewards to preceding memory decisions. Persistent-memory methods such as Memory-R1~\citep{yan2026memoryr1}, Mem-$\alpha$~\citep{wang2025memalpha}, and AtomMem~\citep{huo2026atommem} further extend the learned policy from memory construction and updating toward broader combinations of reading, editing, and answering. More recent work introduces finer-grained supervision to improve credit assignment across long memory-operation sequences. Fine-Mem~\citep{ma2026finemem} and TRUSTMEM~\citep{yang2026trustmem} provide localized feedback for memory construction or updating through intermediate QA signals, downstream usage, or verifier-based evaluation, while keeping retrieval or reasoning largely fixed. Mem-T~\citep{yue2026memt} further connects memory formation, evolution, and multi-step retrieval by evaluating retrieval trajectories and propagating later retrieval outcomes back to earlier construction decisions. These methods move toward process-aware memory optimization, but their process feedback typically targets specific memory stages or estimates the contribution of earlier actions from later outcomes. In contrast, \textbf{AiO-Mem unifies memory interactions as transitions over a shared evidence state and uses a trajectory-conditioned reward model to directly assess the semantic quality of each memory action in a unified reward space, enabling coordinated optimization of retrieval and construction}.

\section{Experiment details}
\label{app:experimental-details}

\subsection{Datasets and baselines}
\label{app:datasets-baselines}

\textbf{Datasets}
\textbf{\ul{The LoCoMo dataset} \citep{maharana2024locomo} \ul{targets the evaluation of long-range conversational memory.}} It features extremely long dialogues, with each conversation spanning roughly 300 turns and around 9K tokens on average. The accompanying questions cover four categories---Single-hop, Multi-hop, Temporal, and Open-domain---capturing different aspects of memory and reasoning in LLMs. Following the evaluation setup of previous work, we report F1 and BLEU-1 (B1) scores for each category as well as the overall performance.

\textbf{\ul{MemoryAgentBench} \citep{hu2026memoryagentbench} \ul{targets the evaluation of memory agents under incremental, multi-turn interactions.}} In our experiments, we evaluate three memory capabilities---Accurate Retrieval, Test-Time Learning, and Long-Range Understanding---using nine tasks from the benchmark. The selected tasks include SH-Doc QA (SH-Doc), MH-Doc QA (MH-Doc), LongMemEval (LME(S)), TREC Coarse (TREC-C), NLU, TREC Fine (TREC-F), CLINC150 (CLINC), BANKING77 (Banking), and $\infty$Bench-Sum (InfBench). 

\textbf{\ul{The PersonaMem benchmark} \citep{jiang2025personamem} \ul{targets the evaluation of dynamic user profiling and personalized response generation over long-term user--LLM interactions.}} Unlike factual memory benchmarks that primarily focus on retrieving explicitly stated information from past interactions, PersonaMem emphasizes user-profile memory, requiring models to infer, maintain, and update users' preferences and underlying traits over time. In our experiments, we evaluate five query types corresponding to five dimensions: Recall User-Shared Facts (Rec.), Generalize to New Scenarios (Gen.), Track Full Preference Evolution (Evol.), Revisit Reasons Behind Preference Updates (Reason), and Provide Preference-Aligned Recommendations (Reco.). Together, these query types assess whether a memory system can recover user-specific facts, generalize learned preferences to new situations, track evolving preferences, reason about the causes of preference changes, and provide recommendations aligned with the user's current profile.

\paragraph{Baselines.}
We compare \textbf{AiO-Mem} against several representative baselines across three groups. 
\circled{1} \textit{Reference:} Long-Context and RAG.
\circled{2} \textit{Agent Memory Harness:} MemGPT~\citep{packer2023memgpt}, MemoryBank~\citep{zhong2024memorybank}, Mem0~\citep{chhikara2025mem0}, MemoryOS~\citep{kang2025memoryos}, LightMem~\citep{fang2026lightmem}, A-Mem~\citep{xu2026amem}, GAM~\citep{yan2025gam}, Zep~\citep{rasmussen2025zep}, LangMem~\citep{langchain2025langmem}, EverMemOS~\citep{hu2026evermemos}, and Emotional RAG~\citep{huang2024emotionalrag}. These approaches equip a frozen backbone model with explicit memory storage, retrieval, summarization, organization,or update mechanisms.
\circled{3} \textit{Trained Method:} MEM1~\citep{zhou2026mem1}, MemAgent~\citep{yu2026memagent}, Mem-T~\citep{yue2026memt}, AtomMem~\citep{huo2026atommem}, Mem-$\alpha$~\citep{wang2025memalpha}, FineMem~\citep{ma2026finemem}, and TrustMem~\citep{yang2026trustmem}. Compared with agent memory harnesses, these baselines either learn memory-oriented agent behaviors or benefit from stronger post-trained reasoning capabilities. To ensure a fair comparison, we follow the same baseline configurations as prior work~\citep{yue2026memt,ma2026finemem}.


\clearpage
\subsection{Implementation Details}
\label{app:implementation-details}
For traceable \textit{multi-view event bank} and \textit{evidence-state transition modeling}, we use Qwen3-8B as the structure-aware compiler $\mathcal{C}$, dense retrieval for the summary and event-phrase views, BM25 for the query-anchor view, and a combination of BM25 and dense retrieval for the raw-turn view. We use BGE-M3\footnote{https://huggingface.co/BAAI/bge-m3.} as the embedding model. We set the maximum interaction horizon to $T_{\max}=8$ and the observation text budget to 10,240 tokens for evidence extraction.

\begin{wrapfigure}{r}{0.5\linewidth}
    \vspace{-0.3cm}
    \centering
    \includegraphics[width=0.9\linewidth]{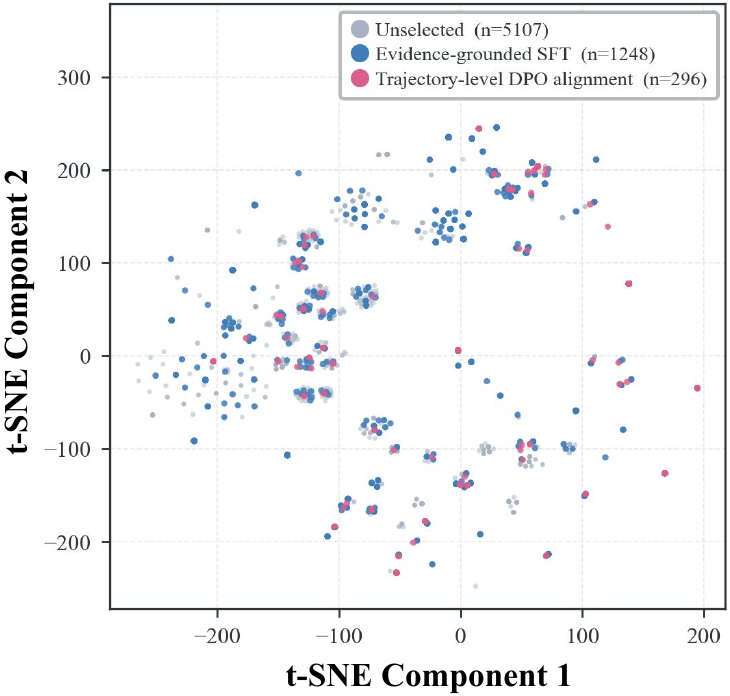}
    \caption{Reward-model trajectories by behavioral region in t-SNE space.}
    \label{fig:rm_data_distribution}
\end{wrapfigure}

For \textit{unified reward modeling}, we construct the training data from policy rollouts. We use Qwen3-8B as the reward model. We filter low-quality and near-duplicate samples using BGE-M3 similarity, then apply agglomerative clustering \citep{johnson1967hierarchical,ward1963hierarchical} to improve diversity and retain at most 10 representative trajectories per question. This yields 1,248 evidence-grounded SFT examples covering 150 questions and 4,162 scoreable steps, with the selected trajectories spanning diverse behavioral regions as shown in Figure~\ref{fig:rm_data_distribution}. We derive action-specific rule scores for retrieval, note construction, citation, and stopping, prioritizing recall of missing evidence over evidence precision. We then fine-tune the reward model with LoRA SFT for 2 epochs using a global batch size of 8, a learning rate of $1\times10^{-4}$, and LoRA rank, alpha, and dropout of 32, 64, and 0.05. For trajectory-level DPO alignment, we use DeepSeek-V4-Flash\footnote{DeepSeek-V4-Flash refers to \texttt{DeepSeek-V4-Flash-0731.}} to review scoreable steps and retain cases differing from the rule-based score by one quality level, yielding 358 preference pairs from 296 trajectories and 102 questions. Human evaluation on 100 sampled steps shows strong agreement with DeepSeek-V4-Flash ($\mathrm{ICC}=0.87$). Finally, the reward model is further optimized with DPO for one epoch using a batch size of 8, a learning rate of $2\times10^{-6}$, and $\beta=0.1$.

For \textit{process-aligned policy optimization}, we use Qwen3-4B as the memory policy actor. We first aggregate step-level score with the action-balanced trajectory reward so that repeated \texttt{search} and \texttt{add} operations do not dominate less frequent but decisive actions such as \texttt{stop}. For reward normalization, we set $r_c^{\mathrm{lo}}=1$ and $r_c^{\mathrm{hi}}=100$ for every $c\in\mathcal{A}$. Following MEM-T~\citep{yue2026memt}, we adopt the same data configuration by splitting the LoCoMo dataset into training, validation, and test sets with a ratio of $1{:}1{:}8$ to ensure a fair comparison. During training, we set the retrieval width to $K_{\mathrm{ret}}=10$, the maximum interaction horizon to $T_{\max}=8$, and the number of rollouts to 8. The policy is optimized with LoRA using a learning rate of $2\times10^{-5}$, with the LoRA rank and alpha set to 32 and 64, respectively. We use Qwen3-4B as the answer model (fine-tuned on our LoCoMo training split for LoCoMo) and enable thinking mode during inference. All experiments are conducted on a single node with 8$\times$ NVIDIA A6000 GPUs (40 GB each), with a training batch size of 8. Detailed prompts are provided in Appendix~\ref{app:prompts}.

\section{Supplementary Experiments}
\label{app:additional-experiments}
\subsection{Sensitivity Analysis}
\label{app:sensitivity}
\begin{table}[!t]
\centering
\caption{Parameter analysis on LoCoMo with different top-$K$ values and maximum interaction steps $T_{\max}$.}
\label{tab:parameter_analysis}
{\fontsize{8pt}{10pt}\selectfont
\renewcommand{\arraystretch}{0.85}

\newlength{\paFirstCol}
\newlength{\paSecondCol}
\newlength{\paTmp}
\newlength{\paSep}

\settowidth{\paFirstCol}{\textbf{top-$K$}}
\settowidth{\paTmp}{$K=15$}
\ifdim\paTmp>\paFirstCol
  \setlength{\paFirstCol}{\paTmp}
\fi
\settowidth{\paSecondCol}{\textbf{Max Step}}

\setlength{\paSep}{\linewidth}
\addtolength{\paSep}{-\paFirstCol}
\addtolength{\paSep}{-\paSecondCol}
\addtolength{\paSep}{-9.2cm}
\divide\paSep by 22
\setlength{\tabcolsep}{\paSep}

\begin{tabular}{@{}ll*{10}{>{\centering\arraybackslash}p{0.92cm}}@{}}
\toprule
\multirow{2}{*}{\textbf{top-$K$}}
& \multirow{2}{*}{\textbf{Max Step}}
& \multicolumn{2}{c}{\textbf{Multi-hop}}
& \multicolumn{2}{c}{\textbf{Temporal}}
& \multicolumn{2}{c}{\textbf{Open-domain}}
& \multicolumn{2}{c}{\textbf{Single-hop}}
& \multicolumn{2}{c}{\textbf{Overall}} \\
\cmidrule(lr){3-4}
\cmidrule(lr){5-6}
\cmidrule(lr){7-8}
\cmidrule(lr){9-10}
\cmidrule(lr){11-12}
&
& F1 & B1
& F1 & B1
& F1 & B1
& F1 & B1
& F1 & B1 \\
\midrule
\multirow{4}{*}{$K=5$}
& $T_{\max}=4$
& 44.6 & 35.7
& 66.2 & 59.6
& 30.3 & 25.6
& 62.7 & 57.3
& 58.0 & 51.8 \\
& $T_{\max}=6$
& 43.5 & 35.4
& 62.4 & 53.3
& 34.9 & 30.9
& 65.9 & 61.5
& 59.2 & 53.2 \\
& $T_{\max}=8$
& 44.0 & 36.6
& 63.9 & 55.7
& 34.4 & 30.3
& 65.4 & 60.9
& 59.2 & 53.5 \\
& \textbf{$T_{\max}=10$}
& 43.0 & 35.1
& 61.9 & 53.1
& 33.5 & 29.3
& 65.8 & 61.1
& 58.8 & 52.8 \\
\midrule
\multirow{4}{*}{$\mathbf{K=10}$}
& $T_{\max}=4$
& 43.6 & 35.6
& 63.4 & 54.6
& 33.7 & 30.2
& 65.9 & 60.9
& 59.3 & 53.3 \\
& $T_{\max}=6$
& 46.9 & 38.1
& 66.5 & 60.6
& 32.4 & 29.6
& 66.1 & 61.2
& 60.7 & 54.9 \\
& $\mathbf{T_{\max}=8}$
& \textbf{47.1} & \textbf{38.8}
& \textbf{66.0} & \textbf{59.4}
& \textbf{33.7} & \textbf{30.5}
& \textbf{67.0} & \textbf{61.5}
& \textbf{61.0} & \textbf{55.0} \\
& $T_{\max}=10$
& 43.5 & 35.3
& 66.1 & 59.2
& 33.0 & 28.9
& 65.9 & 60.1
& 59.8 & 53.4 \\
\midrule
\multirow{4}{*}{$K=15$}
& $T_{\max}=4$
& 43.5 & 35.4
& 61.2 & 52.0
& 32.6 & 29.2
& 66.5 & 61.9
& 59.1 & 53.0 \\
& $T_{\max}=6$
& 43.5 & 35.7
& 63.3 & 53.6
& 34.8 & 30.9
& 65.8 & 61.0
& 59.3 & 53.0 \\
& $T_{\max}=8$
& 45.5 & 36.7
& 65.7 & 60.6
& 31.2 & 28.2
& 66.5 & 60.9
& 60.3 & 54.4 \\
& $T_{\max}=10$
& 45.1 & 36.1
& 65.0 & 58.8
& 30.6 & 26.9
& 65.8 & 60.5
& 59.6 & 53.6 \\
\bottomrule
\end{tabular}
}
\end{table}
Table~\ref{tab:parameter_analysis} reports the detailed F1 and B1 results for all evaluated top-$K$ values and maximum interaction steps $T_{\max}$ on LoCoMo. Models trained with different $K$ and $T_{\max}$ settings consistently achieve strong performance, demonstrating the generalizability of our method across training budgets. Overall F1 is highest at $K=10, T_{\max}=8$, which we adopt as our configuration.

\subsection{Additional Results on PersonaMem}
\label{app:personamem}
\begingroup
\providecommand{\ph}[1]{\textcolor{red}{#1}}
\definecolor{headerblue}{RGB}{234,240,250}
\providecolor{srcblue}{RGB}{18,62,138}
\providecolor{estorange}{RGB}{201,110,17}
\providecommand{\src}[1]{\textcolor{srcblue}{#1}}
\providecommand{\est}[1]{\textcolor{estorange}{#1}}

\definecolor{hltop1}{RGB}{140,190,238}
\definecolor{hltop2}{RGB}{214,233,250}
\newcommand{\hlone}[1]{\cellcolor{hltop1}#1}
\newcommand{\hltwo}[1]{\cellcolor{hltop2}#1}

\begin{center}
\begin{minipage}{0.75\columnwidth}
\centering
\tiny
\captionof{table}{Main results on \textbf{PersonaMem}.}
\label{tab:memory_capability_results}

\renewcommand{\arraystretch}{0.95}

\begin{adjustbox}{width=\linewidth,center}
\begin{tabular}{@{}lcccccc c@{}}
\toprule
\textbf{Method}
& \textbf{Rec.}
& \textbf{Gen.}
& \textbf{Evol.}
& \textbf{Reason}
& \textbf{Reco.}
& \textbf{Overall}
& \\
\midrule

Claude 3.7-Sonnet
& 25.0
& 29.0
& 45.0
& 57.0
& 43.0
& 40.0
& \\
\midrule

Mem0
& 32.1
& 57.9
& 54.7
& 80.8
& 52.7
& 53.4
& \\

Zep
& 35.3
& 42.1
& 34.5
& 50.5
& 36.4
& 39.1
& \\

LangMem
& 31.3
& 8.8
& 53.2
& 81.8
& 40.0
& 46.4
& \\

A-MEM
& \underline{63.0}
& 57.9
& 54.7
& \textbf{85.9}
& \textbf{69.1}
& \underline{65.3}
& \\

MemoryOS
& 47.1
& 56.1
& 32.4
& 48.5
& 47.3
& 44.3
& \\
\midrule

MemoryBank
& 35.3
& 49.1
& 51.1
& 56.6
& 40.0
& 46.5
& \\

EverMemOS
& 47.1
& 47.4
& 41.0
& 43.4
& \underline{56.4}
& 45.7
& \\

Emotional RAG
& 52.9
& \underline{61.4}
& 44.6
& 58.6
& 43.6
& 51.6
& \\
\midrule

Mem-T
& 61.6
& \underline{61.4}
& \underline{56.1}
& 72.7
& 45.5
& 60.5
& \\

AiO-Mem
& \hltwo{\textbf{71.9}}
& \hltwo{\textbf{68.4}}
& \hltwo{\textbf{62.6}}
& \hltwo{\underline{82.8}}
& \hltwo{54.6}
& \hltwo{\textbf{69.2}}
& \\

\bottomrule
\end{tabular}
\end{adjustbox}

\end{minipage}
\end{center}
\endgroup

We further conduct experiments on PersonaMem \citep{jiang2025personamem}. It evaluates the ability of memory systems to maintain and use dynamic user profiles over long-term interactions. As shown in Table~\ref{tab:memory_capability_results}, AiO-Mem achieves the best overall performance with an overall score of 0.692. In contrast, existing memory systems show less balanced performance across different capabilities. A-MEM performs well on reasoning and recommendation but is weaker on generalization and preference evolution. Mem-T achieves competitive performance on recall and generalization but performs less effectively on preference-aligned recommendation. These results suggest that \textbf{the unified memory interaction in AiO-Mem enables effective retrieval, organization, and use of user-profile information from long-term interaction histories.}


\subsection{Multi-View Retrieval Complementarity Study}
\label{app:multiview-complementarity}

\begin{figure*}[t]
    \centering

    \begin{minipage}[b]{0.52\linewidth}
        \centering
        \begin{subfigure}[b]{\linewidth}
            \centering
            \includegraphics[width=\linewidth]{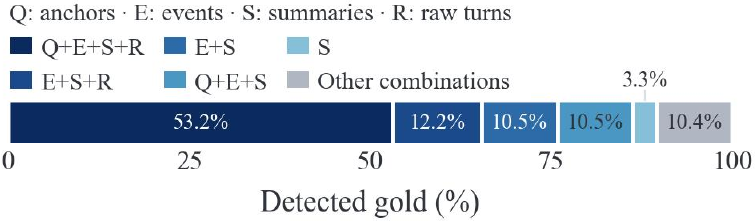}
            \caption{Major view combinations}
            \label{fig:view-overview}
        \end{subfigure}

        \begin{subfigure}[b]{\linewidth}
            \centering
            \includegraphics[width=\linewidth]{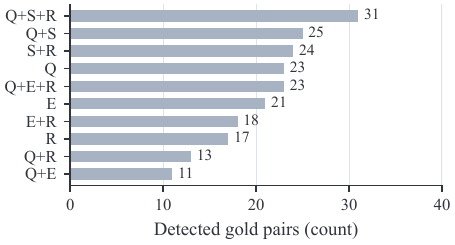}
            \caption{Remaining view combinations}
            \label{fig:view-tail}
        \end{subfigure}
    \end{minipage}
    \hfill
    \begin{minipage}[b]{0.46\linewidth}
        \centering
        \begin{subfigure}[b]{\linewidth}
            \centering
            \includegraphics[width=\linewidth]{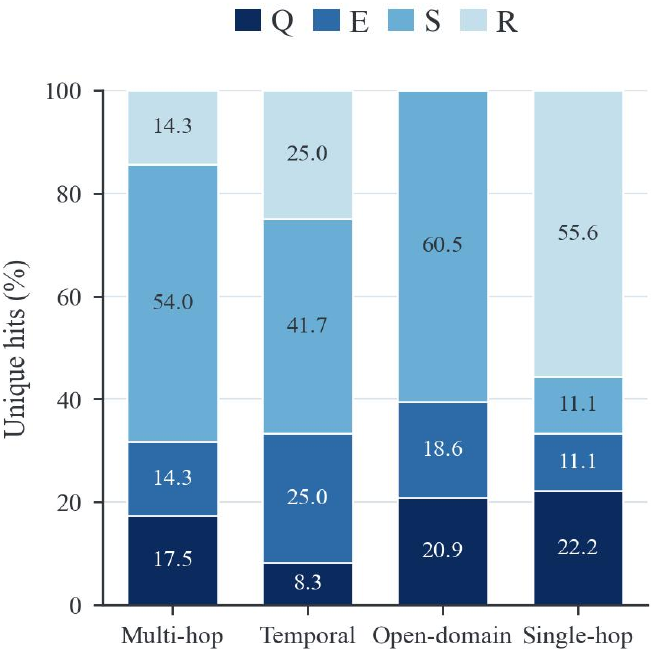}
            \caption{Single-view hits by category}
            \label{fig:view-unique-hits}
        \end{subfigure}
    \end{minipage}

    \caption{Complementary evidence coverage across retrieval views.
    Panel~\subref{fig:view-overview} summarizes the major view combinations,
    panel~\subref{fig:view-tail} expands the remaining combinations, and
    panel~\subref{fig:view-unique-hits} shows the sources of evidence detected
    by exactly one view across question categories.}
    \label{fig:multiview-complementarity}
\end{figure*}

We further analyze the complementarity of the four retrieval views using queries generated during actual inference trajectories. After merging candidates across all views, 1,990 of 2,051 question--evidence pairs are detected by at least one view, yielding 97.03\% candidate-pool coverage. As shown in Figures~\ref{fig:view-overview} and \ref{fig:view-tail}, although evidence detected by all four views forms the largest group, substantial evidence is distributed across other view combinations, demonstrating their complementary coverage. Figure~\ref{fig:view-unique-hits} further shows that single-view hits vary across question categories; for example, raw turns and query anchors contribute relatively more to Single-hop questions. These results indicate that different views capture distinct evidence patterns and jointly improve coverage across diverse question types.

\begin{figure}[t]
    \centering
    \includegraphics[width=\linewidth]{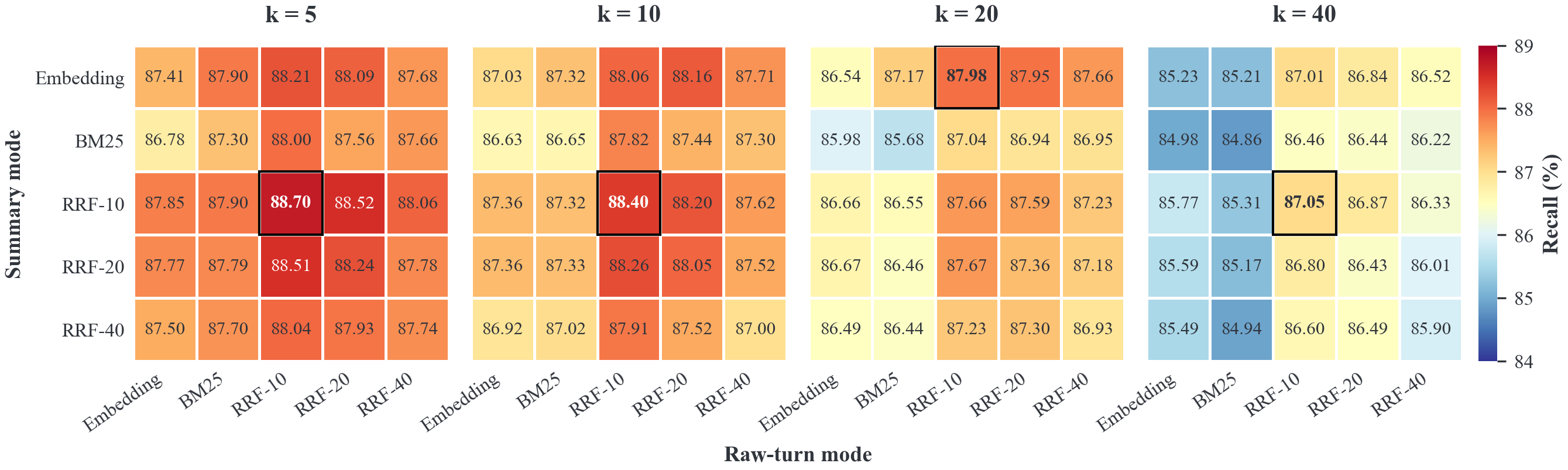}
    \caption{Sensitivity to retrieval modes and RRF parameters. Rows
    and columns denote the summary and raw-turn retrieval modes,
    respectively, while each panel uses a different aggregation RRF
    parameter. Red and blue indicate higher and lower evidence Recall.}
    \label{fig:retrieval-sensitivity}
\end{figure}

\clearpage
\subsection{Multi-View Retrieval Sensitivity Study}
\label{app:inference-retrieval-sensitivity}

\begin{wrapfigure}{r}{0.4\linewidth}
    \vspace{-0.3cm}
    \centering
    \includegraphics[width=\linewidth]{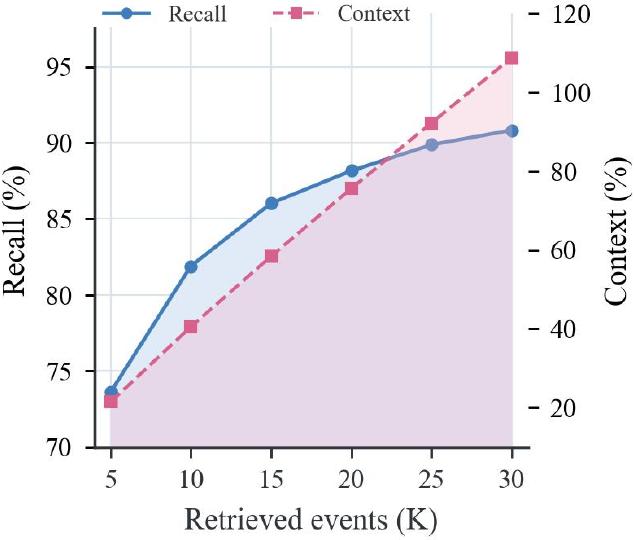}
    \caption{Trade-off between retrieval Recall and context cost.}
    \label{fig:retrieval-budget}
    \vspace{-0.3cm}
\end{wrapfigure}

The retrieval-budget analysis in Figure~\ref{fig:retrieval-budget} characterizes the trade-off between evidence coverage and context cost, while Figure~\ref{fig:retrieval-sensitivity} examines sensitivity to retrieval configurations. As $K_{\mathrm{ret}}$ increases from 5 to 30, evidence Recall improves steadily from 73.62\% to 90.82\%, while the average retrieved context grows substantially from 2,197 to 11,137 tokens. At $K_{\mathrm{ret}}=20$, the system already achieves 88.16\% Recall with 7,743 tokens; increasing the budget to 30 brings only a further 2.66-point Recall gain at approximately 44\% higher context cost, indicating diminishing returns from larger retrieval budgets. We therefore set $K_{\mathrm{ret}}=20$ as a practical balance between evidence coverage and context efficiency. Figure~\ref{fig:retrieval-sensitivity} further shows that moderate variations around the default retrieval modes and RRF parameters cause only limited changes in Recall, whereas larger deviations can lead to noticeable degradation. This suggests that the adopted retrieval configuration lies in a relatively stable and high-performing region rather than depending on a narrowly tuned parameter setting.

\section{Reward and Optimization Analysis}
\label{app:reward-optimization}

\subsection{Unified Reward Model Alignment Study}
\label{app:reward-alignment}

\begin{figure*}[t]
    \centering

    {
        \captionsetup{
            type=table,
            width=\linewidth,
            justification=centering,
            singlelinecheck=false
        }
        \captionof{table}{
            Correlations between Reward scores and evidence metrics.
        }
        \label{tab:reward-alignment}
    }

    \begin{minipage}{0.8\textwidth}
        \centering

        \small
        \setlength{\tabcolsep}{4.0pt}
        \begin{tabular}{lcccc}
            \toprule
            & \multicolumn{2}{c}{Recall}
            & \multicolumn{2}{c}{Precision} \\
            \cmidrule(lr){2-3}
            \cmidrule(lr){4-5}
            Reward model
            & Pearson & Spearman
            & Pearson & Spearman \\
            \midrule

            Qwen3-8B
            & 0.877 & 0.783
            & 0.640 & 0.675 \\

            AiO Reward Model (Ours)
            & \textbf{0.958} & \textbf{0.800}
            & \textbf{0.686} & \textbf{0.861} \\

            \bottomrule
        \end{tabular}
    \end{minipage}

    \vspace{8pt}

    \begin{minipage}{0.8\textwidth}
        \centering

        \begin{subfigure}[t]{0.4\linewidth}
            \centering
            \includegraphics[width=\linewidth]
            {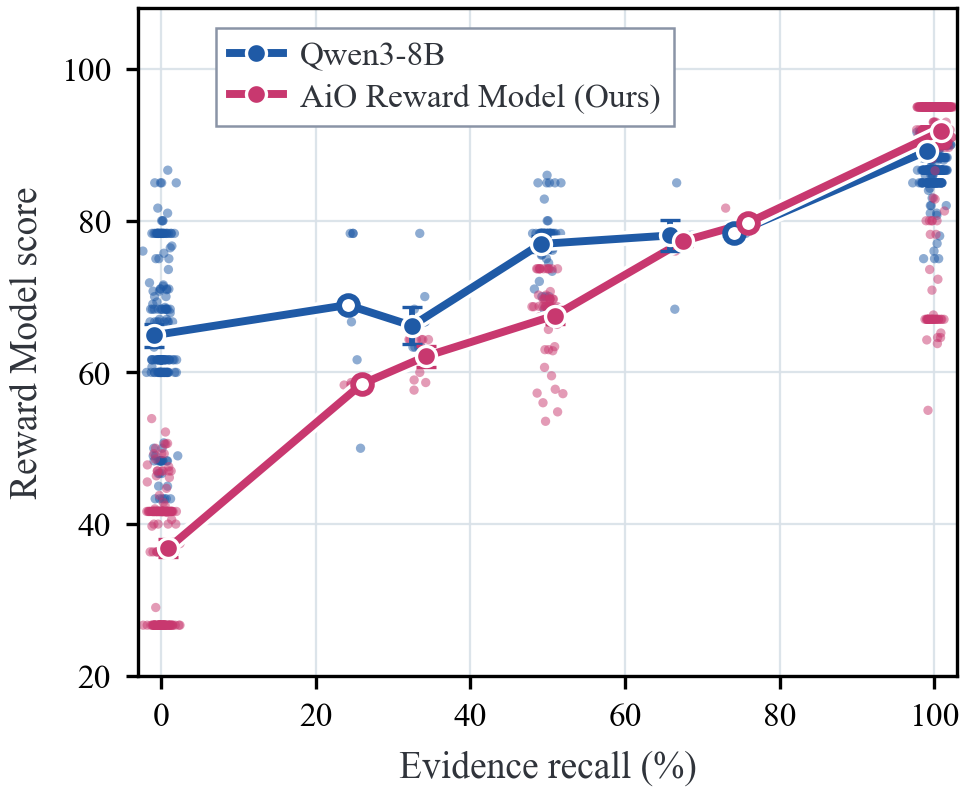}
            \captionsetup{justification=centering}
            \caption{Reward vs. Recall}
            \label{fig:reward-recall-scatter}
        \end{subfigure}
        \hspace{0.025\linewidth}
        \begin{subfigure}[t]{0.4\linewidth}
            \centering
            \includegraphics[width=\linewidth]
            {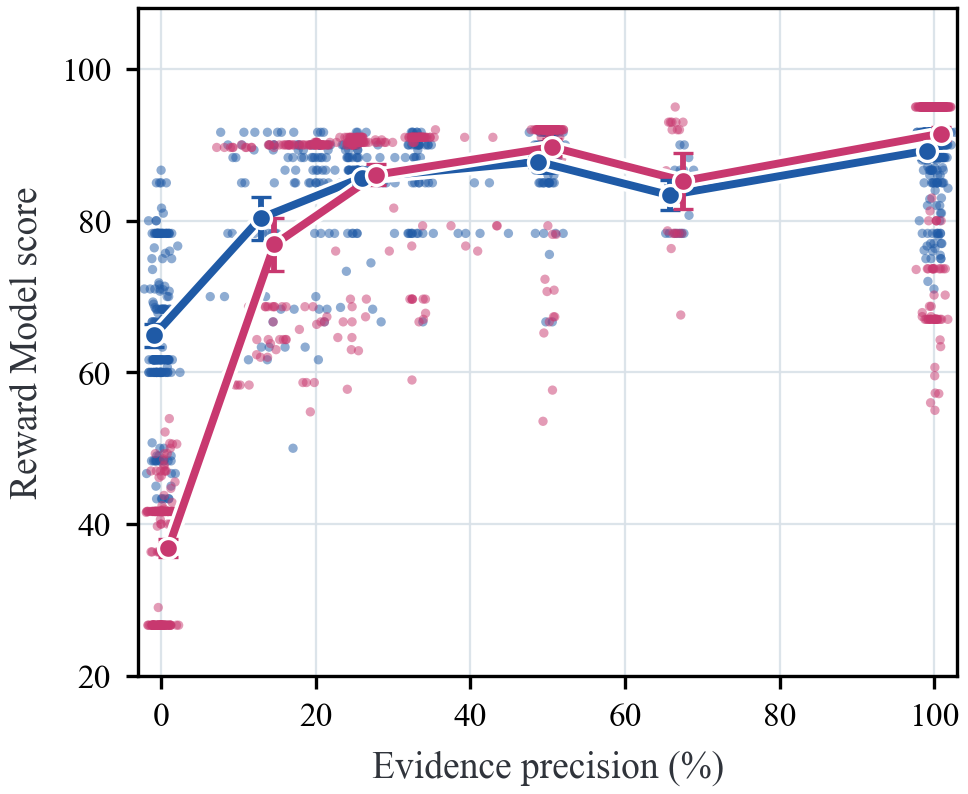}
            \captionsetup{justification=centering}
            \caption{Reward vs. Precision}
            \label{fig:reward-precision-scatter}
        \end{subfigure}

    \end{minipage}

    \captionsetup{
        type=figure,
        width=\linewidth,
        justification=centering,
        singlelinecheck=false
    }
    \caption{
        Sample-level Reward scores versus evidence metrics.
    }
    \label{fig:reward-scatter}

\end{figure*}

We further evaluate whether the trained reward model aligns with evidence usage by measuring its correlation with trajectory-level Recall and Precision. As shown in Table~\ref{tab:reward-alignment}, training increases the Pearson correlation with Recall from 0.877 to 0.958 and the Spearman correlation with Precision from 0.675 to 0.861, indicating stronger alignment with evidence-based trajectory quality. Figures~\ref{fig:reward-recall-scatter} and \ref{fig:reward-precision-scatter} further show that the base reward model assigns relatively concentrated scores across different evidence qualities, whereas the trained model exhibits clearer positive trends with both Recall and Precision. Notably, some trajectories still receive high rewards at different Precision levels, suggesting that the trained reward model captures broader trajectory quality beyond evidence Precision alone.

\subsection{Trajectory-level DPO alignment Study}
\label{app:reward-case-study}

Evidence-grounded SFT provides action-level supervision based on evidence coverage and provenance, but such rule-based signals cannot fully assess the semantic quality of an action within its trajectory context. As introduced in Section~\ref{sec:reward}, trajectory-level DPO addresses this limitation by using full trajectory context to align the reward model with finer-grained semantic judgments. To visualize how trajectory-level DPO distinguishes trajectories beyond rule-based evidence matching, we compare four matched trajectory pairs for the same questions. Within each pair, the trajectories have identical Evidence-ID Recall and Precision but receive different scores after trajectory-level DPO alignment.

Specifically, in q114, both trajectories cite the correct evidence ID, but the lower-Reward trajectory mixes visual details from another event into the adoption meeting, whereas the higher-Reward trajectory remains focused on the target event. In q76, both obtain zero Evidence-ID Recall and Precision, yet the higher-Reward trajectory retains evidence supporting ``three children,'' while the lower-Reward one infers ``two children'' from insufficient observations. In q65, despite complete evidence coverage, the lower-Reward trajectory repeats the same fact and conflates Melanie's stated preference with the inference about Vivaldi, whereas the higher-Reward trajectory separates the source fact from the inference. In q7, the lower-Reward trajectory mixes the target future camping plan with a past camping event, while the higher-Reward trajectory preserves the correct temporal focus. These cases show that trajectory-level DPO complements rule-based supervision by aligning the reward model with evidence attribution, relevance, factual organization, temporal consistency, and redundancy beyond Evidence-ID matching.

\clearpage
\begingroup
\definecolor{AiOCaseBlue}{HTML}{2E5AA8}
\definecolor{AiOCaseRed}{HTML}{E60033}
\definecolor{AiOCaseTeal}{HTML}{008080}

\tcbset{
  AiORewardCase/.style={
    colback=AiOCaseBlue!10!white,
    colframe=AiOCaseBlue,
    coltext=black,
    width=\linewidth,
    arc=5mm,
    boxrule=0.8pt,
    left=9pt,right=9pt,top=8pt,bottom=8pt,
    before skip=10pt,after skip=10pt,
    fontupper=\small,
    before upper={\setlength{\parindent}{0pt}\setlength{\parskip}{0pt}}
  }
}

\long\def\AiOScoreHeading#1#2#3{%
  \par\medskip\noindent
  \textcolor{#1}{\textbf{#2 Trajectory-level Reward Model Score}%
  \hfill\textbf{Score: #3}}\par
  \nobreak\vspace{3pt}
  {\color{#1}\hrule height 0.4pt}%
  \nobreak\vspace{5pt}
}
\long\def\AiOCaseAnnotation#1#2{%
  \noindent\textcolor{#1}{\textit{Case annotation.} #2}\par
}
\long\def\AiORewardNote#1#2#3{%
  \item\textcolor{#1}{#2 \textit{(Evidence: \texttt{#3})}}%
}

\begin{tcolorbox}[AiORewardCase]
\textbf{Matched Trajectories for q114}
\par\vspace{4pt}
\textbf{Question:} What did Caroline see at the council meeting for adoption?

\par\smallskip
\textbf{Gold answer:} Many people wanting to create loving homes for children in need.

\par\smallskip
\textbf{Gold evidence:} \texttt{D8:9}

\par\smallskip
\textbf{Both trajectories:}
Evidence-ID Recall = \textbf{1.00};
Precision = \textbf{1.00}.

\AiOScoreHeading{AiOCaseRed}{Lower}{0.67}
\AiOCaseAnnotation{AiOCaseRed}{The notes retain the relevant observation about prospective adoptive families, but also attach visual details from another event to the same meeting. Both statements cite the gold evidence ID, concealing the mismatch between the cited source and the additional claim.}

\begin{itemize}[
  label=\textcolor{black}{\textbullet},
  leftmargin=1.5em,labelsep=0.5em,
  topsep=4pt,itemsep=4pt,parsep=0pt,partopsep=0pt
]
  \AiORewardNote{AiOCaseRed}
  {Caroline described the adoption council meeting as inspiring and emotional, noting that many people wanted to create loving homes for children in need.}
  {D8:9}
  \AiORewardNote{AiOCaseRed}
  {Caroline saw \textbf{a sign for a new arrival and an information and domestic building} during the adoption council meeting.}
  {D8:9}

\end{itemize}

\AiOScoreHeading{AiOCaseTeal}{Higher}{0.95}
\AiOCaseAnnotation{AiOCaseTeal}{The notes consistently describe the adoption meeting and the people seeking to provide loving homes. Although the two entries overlap, both remain focused on what Caroline observed at the target event, without introducing the unrelated visual details.}

\begin{itemize}[
  label=\textcolor{black}{\textbullet},
  leftmargin=1.5em,labelsep=0.5em,
  topsep=4pt,itemsep=4pt,parsep=0pt,partopsep=0pt
]
  \AiORewardNote{AiOCaseTeal}
  {Caroline attended an adoption council meeting and saw \textbf{people seeking to create loving homes for children in need}.}
  {D8:9}
  \AiORewardNote{AiOCaseTeal}
  {Caroline described the adoption council meeting as inspiring and emotional, with attendees focused on \textbf{creating loving homes for children in need}.}
  {D8:9}
\end{itemize}
\end{tcolorbox}

\begin{tcolorbox}[AiORewardCase]
\textbf{Matched Trajectories for q76}
\par\vspace{4pt}
\textbf{Question:} How many children does Melanie have?

\par\smallskip
\textbf{Gold answer:} Three.

\par\smallskip
\textbf{Gold evidence:} \texttt{D18:1, D18:7}

\par\smallskip
\textbf{Both trajectories:}
Evidence-ID Recall = \textbf{0.00};
Precision = \textbf{0.00}.

\AiOScoreHeading{AiOCaseRed}{Lower}{0.27}
\AiOCaseAnnotation{AiOCaseRed}{The notes use observations of two children in particular scenes to support a claim about Melanie's total number of children. Seeing two children at a campfire or museum does not establish that she has only two, leaving the numerical conclusion insufficiently supported.}

\begin{itemize}[
  label=\textcolor{black}{\textbullet},
  leftmargin=1.5em,labelsep=0.5em,
  topsep=4pt,itemsep=4pt,parsep=0pt,partopsep=0pt
]
  \AiORewardNote{AiOCaseRed}
  {Melanie \textbf{has two children based on the image of a man and two children} in the campfire scene.}
  {D8:32}
  \AiORewardNote{AiOCaseRed}
  {Melanie took two children to the museum.}
  {D6:4}

\end{itemize}

\AiOScoreHeading{AiOCaseTeal}{Higher}{0.53}
\AiOCaseAnnotation{AiOCaseTeal}{The notes preserve an observation involving three children and additional family context. Neither entry matches the annotated gold evidence IDs, but the beach observation is more relevant to the reference answer. The displayed notes alone still do not conclusively establish the total number of Melanie's children.}

\begin{itemize}[
  label=\textcolor{black}{\textbullet},
  leftmargin=1.5em,labelsep=0.5em,
  topsep=4pt,itemsep=4pt,parsep=0pt,partopsep=0pt
]
  \AiORewardNote{AiOCaseTeal}
  {Melanie went to the beach with \textbf{three children}.}
  {D10:8}
  \AiORewardNote{AiOCaseTeal}
  {Melanie's children were at the Grand Canyon with her.}
  {D18:5}
\end{itemize}
\end{tcolorbox}

\begin{tcolorbox}[AiORewardCase]
\textbf{Matched Trajectories for q65}
\par\vspace{4pt}
\textbf{Question:} Would Melanie likely enjoy the song ``The Four Seasons'' by Vivaldi?

\par\smallskip
\textbf{Gold answer:} Yes; it is classical music.

\par\smallskip
\textbf{Gold evidence:} \texttt{D15:28}

\par\smallskip
\textbf{Both trajectories:}
Evidence-ID Recall = \textbf{1.00};
Precision = \textbf{1.00}.

\AiOScoreHeading{AiOCaseRed}{Lower}{0.55}
\AiOCaseAnnotation{AiOCaseRed}{The notes repeat Melanie's classical music preference and conflate it with an inferred preference for Vivaldi.}

\begin{itemize}[
  label=\textcolor{black}{\textbullet},
  leftmargin=1.5em,labelsep=0.5em,
  topsep=4pt,itemsep=4pt,parsep=0pt,partopsep=0pt
]
  \AiORewardNote{AiOCaseRed}
  {Melanie enjoys classical music including composers like Bach and Mozart.}
  {D15:28}
  \AiORewardNote{AiOCaseRed}
  {\textbf{Melanie enjoys classical music including composers like Bach and Mozart.}}
  {D15:28}
  \AiORewardNote{AiOCaseRed}
  {Melanie appreciates classical music genres.}
  {D15:28}
  \AiORewardNote{AiOCaseRed}
  {Melanie appreciates classical music genres including composers like Bach, Mozart, and \textbf{implicitly Vivaldi}.}
  {D15:28}

\end{itemize}

\AiOScoreHeading{AiOCaseTeal}{Higher}{0.95}
\AiOCaseAnnotation{AiOCaseTeal}{The notes distinguish Melanie's stated preference, the genre connection, and the inference that she would likely enjoy Vivaldi.}

\begin{itemize}[
  label=\textcolor{black}{\textbullet},
  leftmargin=1.5em,labelsep=0.5em,
  topsep=4pt,itemsep=4pt,parsep=0pt,partopsep=0pt
]
  \AiORewardNote{AiOCaseTeal}
  {Melanie enjoys \textbf{classical music by Bach and Mozart}, as well as modern music like Ed Sheeran's ``Perfect''.}
  {D15:28}
  \AiORewardNote{AiOCaseTeal}
  {Vivaldi's ``The Four Seasons'' is \textbf{a classical composition}, and Melanie appreciates classical music by Bach and Mozart.}
  {D15:28}
  \AiORewardNote{AiOCaseTeal}
  {Melanie's appreciation for classical music \textbf{suggests she would likely enjoy} Vivaldi's ``The Four Seasons''.}
  {D15:28}
\end{itemize}
\end{tcolorbox}

\begin{tcolorbox}[AiORewardCase]
\textbf{Matched Trajectories for q7}
\par\vspace{4pt}
\textbf{Question:} When is Melanie planning on going camping?

\par\smallskip
\textbf{Gold answer:} June 2023.

\par\smallskip
\textbf{Gold evidence:} \texttt{D2:7}

\par\smallskip
\textbf{Both trajectories:}
Evidence-ID Recall = \textbf{1.00};
Precision = \textbf{0.50}.

\AiOScoreHeading{AiOCaseRed}{Lower}{0.54}
\AiOCaseAnnotation{AiOCaseRed}{The notes repeatedly alternate between a future camping plan and a separate camping trip that had already occurred. Although the relevant expression ``next month'' eventually appears, repetition and the interleaving of different events make the target plan less prominent.}

\begin{itemize}[
  label=\textcolor{black}{\textbullet},
  leftmargin=1.5em,labelsep=0.5em,
  topsep=4pt,itemsep=4pt,parsep=0pt,partopsep=0pt
]
  \AiORewardNote{AiOCaseRed}
  {Melanie plans a camping trip for summer break.}
  {D2:7}
  \AiORewardNote{AiOCaseRed}
  {Melanie \textbf{went camping with kids a few weeks ago}.}
  {D16:2}
  \AiORewardNote{AiOCaseRed}
  {\textbf{Melanie plans a camping trip for summer break.}}
  {D2:7}
  \AiORewardNote{AiOCaseRed}
  {Melanie \textbf{went camping with kids a few weeks ago}.}
  {D16:2}
  \AiORewardNote{AiOCaseRed}
  {Melanie plans a camping trip for next month during summer break.}
  {D2:7}

\end{itemize}

\AiOScoreHeading{AiOCaseTeal}{Higher}{0.92}
\AiOCaseAnnotation{AiOCaseTeal}{The notes foreground the upcoming trip and its time expression, ``next month,'' followed by broader summer-outing context. They avoid repeatedly mixing the plan with the completed camping event, making the information relevant to the temporal question more accessible.}

\begin{itemize}[
  label=\textcolor{black}{\textbullet},
  leftmargin=1.5em,labelsep=0.5em,
  topsep=4pt,itemsep=4pt,parsep=0pt,partopsep=0pt
]
  \AiORewardNote{AiOCaseTeal}
  {Melanie mentioned \textbf{planning a camping trip for next month} during summer break.}
  {D2:7}
  \AiORewardNote{AiOCaseTeal}
  {Melanie agreed to plan a family outing this summer, which includes camping activities.}
  {D12:18}
\end{itemize}
\end{tcolorbox}

\endgroup

\subsection{Training Dynamics}
\label{app:reward-dynamics}

\begin{figure*}[t]
    \centering

    \begin{minipage}{0.78\textwidth}
        \centering

        \begin{subfigure}[t]{0.46\linewidth}
            \centering
            \includegraphics[width=\linewidth]{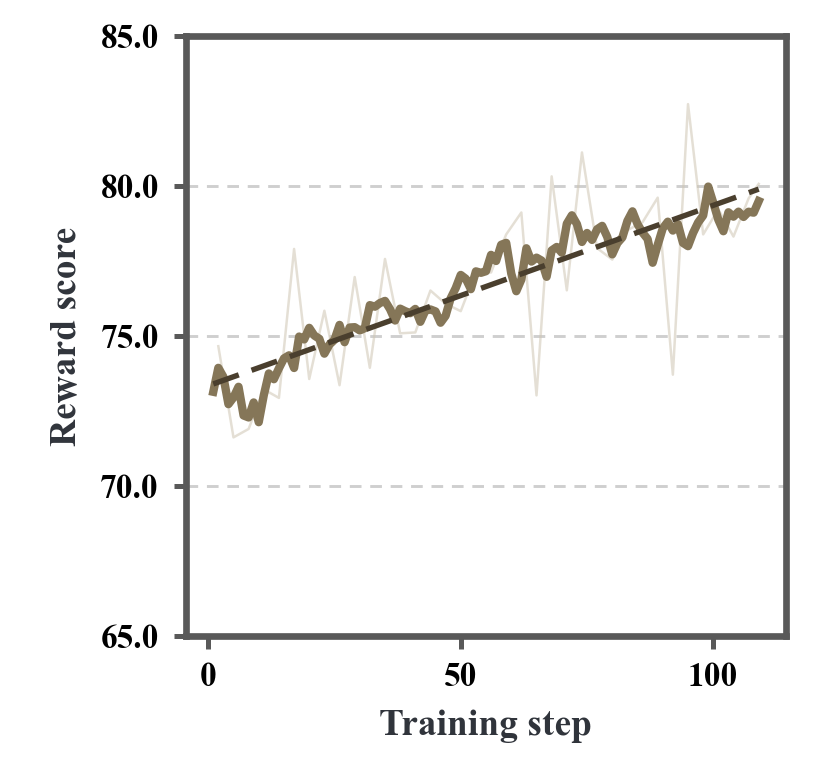}
            \caption{Overall Reward}
            \label{fig:dapo-overall-reward}
        \end{subfigure}
        \hfill
        \begin{subfigure}[t]{0.46\linewidth}
            \centering
            \includegraphics[width=\linewidth]{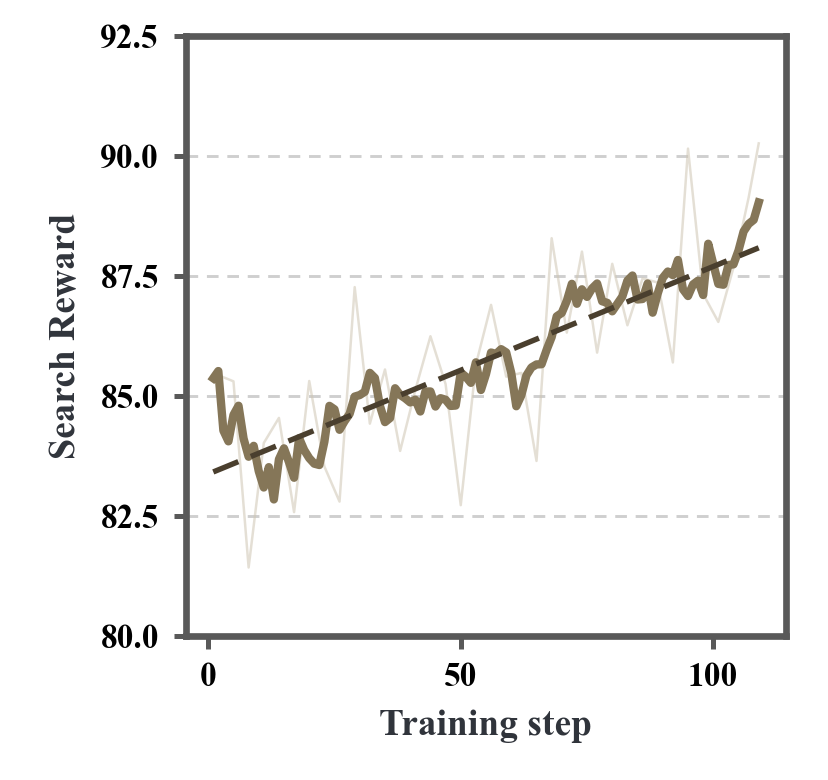}
            \caption{Search Reward}
            \label{fig:dapo-search-reward}
        \end{subfigure}

        \vspace{0.6em}

        \begin{subfigure}[t]{0.46\linewidth}
            \centering
            \includegraphics[width=\linewidth]{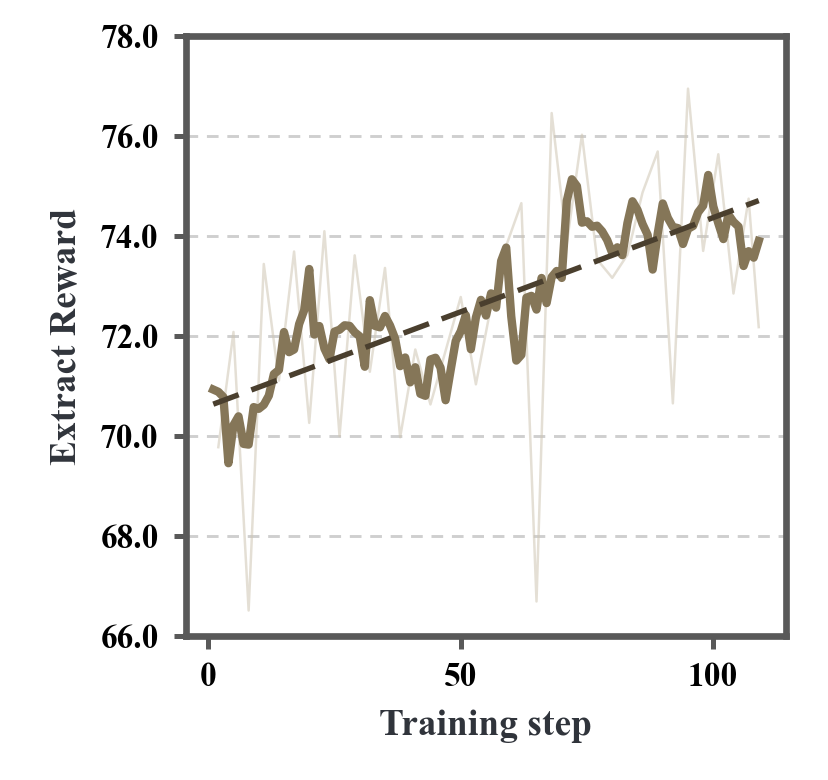}
            \caption{Extract Reward}
            \label{fig:dapo-extract-reward}
        \end{subfigure}
        \hfill
        \begin{subfigure}[t]{0.46\linewidth}
            \centering
            \includegraphics[width=\linewidth]{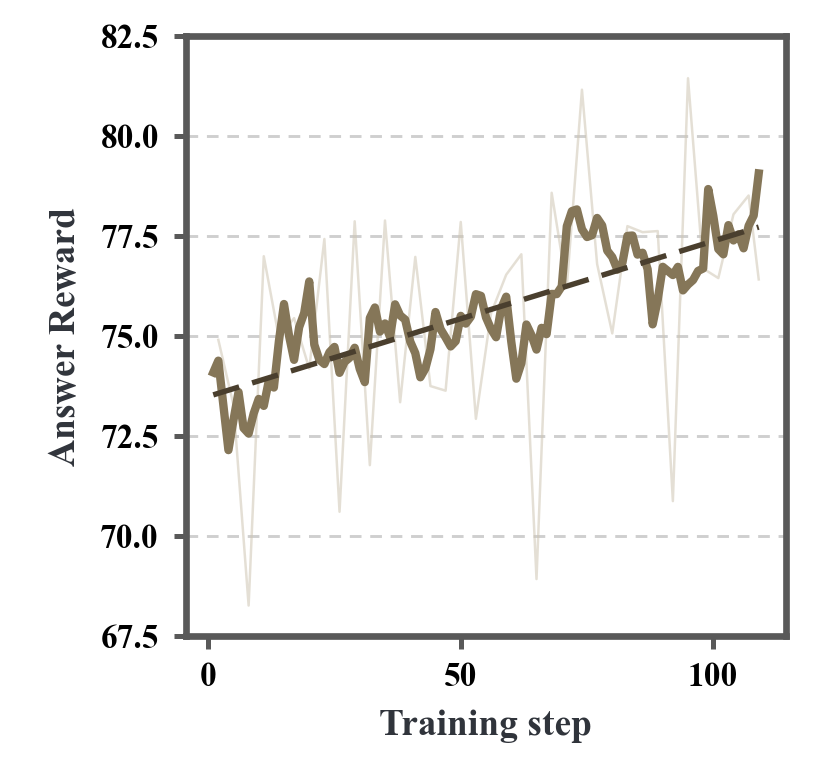}
            \caption{Answer-Decision Reward}
            \label{fig:dapo-answer-decision-reward}
        \end{subfigure}

    \end{minipage}

    \caption{
        Training dynamics.
    }
    \label{fig:dapo-reward-dynamics}
\end{figure*}

Figure~\ref{fig:dapo-reward-dynamics} presents the Reward dynamics during
PA-DAPO training. As shown in Figure~\ref{fig:dapo-overall-reward}, the Overall
Reward exhibits a clear upward trend as training progresses. Although the
Reward varies across individual training steps, the smoothed curve increases
overall, indicating a gradual improvement in the quality of the sampled
trajectories.

Figures~\ref{fig:dapo-search-reward}--\ref{fig:dapo-answer-decision-reward} further decompose the Overall Reward by operation: Search corresponds to search, Extract includes add and skip, and Answer-Decision corresponds to stop.The Search Reward shows an overall increase,
indicating improved evidence retrieval behavior. The Extract Reward also
increases over training, reflecting improvements in evidence extraction and
organization. The Answer-Decision Reward exhibits a similar upward trend,
indicating that the model progressively learns when to stop retrieving and
proceed to answer generation.

The overall improvement across all three stages indicates that the increase
in Overall Reward is not driven by a single interaction stage. Instead, it
reflects joint improvements in evidence retrieval, evidence extraction and
organization, and the decision to stop retrieving and generate an answer.

\subsection{Process-Aligned Evidence Flow Study}
\label{app:evidence-flow}

\begin{figure*}[t]
    \centering
    \begin{subfigure}[t]{0.32\linewidth}
        \centering
        \includegraphics[width=\linewidth]{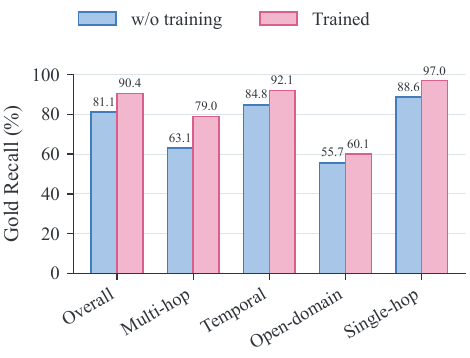}
        \caption{Search}
        \label{fig:evidence-flow-search}
    \end{subfigure}
    \hfill
    \begin{subfigure}[t]{0.32\linewidth}
        \centering
        \includegraphics[width=\linewidth]{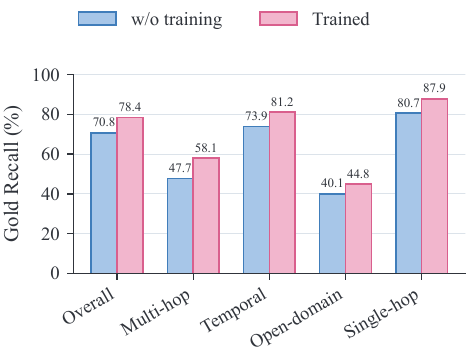}
        \caption{Final working memory}
        \label{fig:evidence-flow-note}
    \end{subfigure}
    \hfill
    \begin{subfigure}[t]{0.32\linewidth}
        \centering
        \includegraphics[width=\linewidth]{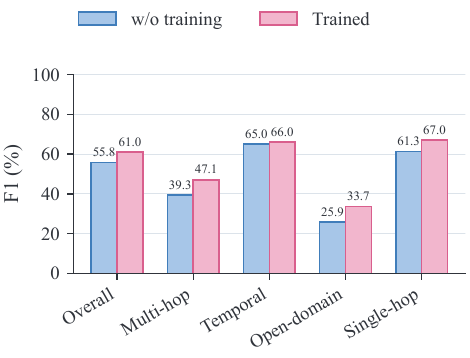}
        \caption{Answer}
        \label{fig:evidence-flow-answer}
    \end{subfigure}
    \caption{Effects of policy training on evidence flow through the memory
    pipeline. The three panels report Search Recall, Final working memory Recall, and
    Answer F1, respectively.}
    \label{fig:evidence-flow}
\end{figure*}

We first analyze how training affects evidence flow through the memory
pipeline. Figures~\ref{fig:evidence-flow-search},
\ref{fig:evidence-flow-note}, and \ref{fig:evidence-flow-answer} compare
the trained and untrained systems at the Search, Final working memory, and Answer
stages, respectively, with further breakdowns by question category.

After training, the system covers more gold evidence during Search and
retains more evidence in the Final working memory, with improvements across all
question categories. The final Answer F1 also increases accordingly.
These results show that policy training improves the entire information
processing pipeline, from evidence retrieval and organization to answer
generation.

\section{Computational Complexity}
\label{app:complexity}

\begin{table}[ht]
\small
\centering
\caption{Complexity comparison between AiO-Mem and OMS.}
\label{tab:aio-complexity}
\setlength{\tabcolsep}{3.5pt}
\resizebox{\linewidth}{!}{%
\begin{tabular}{@{}lccc@{}}
\toprule
\textbf{System}
& \textbf{(I) Segment \& Retrieval}
& \textbf{(II) Construction}
& \textbf{(III) Inference} \\
\midrule

OMS
&
\begin{tabular}[c]{@{}c@{}}
$O(N)\cdot\mathcal{C}_{\mathrm{seg}}$ \\
$+\,O(N\ell)\cdot\mathcal{C}_{\mathrm{LLM}}$ \\
$+\,O(NM_1)\cdot\mathcal{C}_{\mathrm{emb}}$
\end{tabular}
&
\begin{tabular}[c]{@{}c@{}}
$O(NM_1)\cdot\mathcal{C}_{\mathrm{sim}}$ \\
$+\,O(NM_1R_1)\cdot
(\mathcal{C}_{\mathrm{LLM}}+\mathcal{C}_{\mathrm{emb}})$
\end{tabular}
&
\begin{tabular}[c]{@{}c@{}}
$O(N\ell+NM_1R_1)\cdot\mathcal{C}_{\mathrm{LLM}}$ \\
$+\,O(NM_1(1+R_1))\cdot\mathcal{C}_{\mathrm{emb}}$ \\
$+\,O(NM_1)\cdot\mathcal{C}_{\mathrm{sim}}$
\end{tabular}
\\

\midrule

AiO-Mem
&
\begin{tabular}[c]{@{}c@{}}
$O(T_{\max}M)\cdot\mathcal{C}_{\mathrm{sim}}$ \\
$+\,O(N\ell)\cdot\mathcal{C}_{\mathrm{LLM}}$ \\
$+\,O(M+T_{\max})\cdot\mathcal{C}_{\mathrm{emb}}$
\end{tabular}
&
\begin{tabular}[c]{@{}c@{}}
$O(T_{\max}K_{\mathrm{ret}}\ell_e)
\cdot\mathcal{C}_{\mathrm{LLM}}$
\end{tabular}
&
\begin{tabular}[c]{@{}c@{}}
$O(N\ell+T_{\max}K_{\mathrm{ret}}\ell_e)
\cdot\mathcal{C}_{\mathrm{LLM}}$ \\
$+\,O(M+T_{\max})\cdot\mathcal{C}_{\mathrm{emb}}$ \\
$+\,O(T_{\max}M)\cdot\mathcal{C}_{\mathrm{sim}}$
\end{tabular}
\\

\bottomrule
\end{tabular}%
}
\end{table}

As shown in Table~\ref{tab:aio-complexity}, we analyze a long-horizon
history $\mathcal{H}=(h_1,\ldots,h_N)$ consisting of $N$ source units,
each containing an average of $\ell$ tokens, and evaluate the cost of
answering a memory query $q$. In other
memory systems (OMS), the history is first segmented into source units, where each unit
requires an LLM-based summarization and is further encoded into $M_1$
persistent memory entries:
\begin{equation}
\mathcal{C}^{\mathrm{I}}_{\mathrm{OMS}}
=
O(N)\cdot\mathcal{C}_{\mathrm{seg}}
+
O(N\ell)\cdot\mathcal{C}_{\mathrm{LLM}}
+
O(NM_1)\cdot\mathcal{C}_{\mathrm{emb}} .
\end{equation}

During memory construction, the resulting $NM_1$ entries are compared with
the existing memory. A fraction $R_1$ of these entries triggers an
LLM-based rewrite, followed by re-embedding for index update:
\begin{equation}
\mathcal{C}^{\mathrm{II}}_{\mathrm{OMS}}
=
O(NM_1)\cdot\mathcal{C}_{\mathrm{sim}}
+
O(NM_1R_1)\cdot
\left(
\mathcal{C}_{\mathrm{LLM}}
+
\mathcal{C}_{\mathrm{emb}}
\right).
\end{equation}

This results in a total inference cost of:
\begin{equation}
\begin{aligned}
\mathcal{C}^{\mathrm{infer}}_{\mathrm{OMS}}
={}&
O(N\ell+NM_1R_1)\cdot\mathcal{C}_{\mathrm{LLM}}
\\
&+
O(NM_1(1+R_1))\cdot\mathcal{C}_{\mathrm{emb}}
+
O(NM_1)\cdot\mathcal{C}_{\mathrm{sim}} .
\end{aligned}
\end{equation}
Thus, both the LLM-based summarization and memory rewriting costs increase
with the history size and the number of accumulated memory entries.

In \textbf{AiO-Mem}, the history is first compiled into the traceable event bank
$\mathcal{B}=\mathcal{C}(\mathcal{H})$. The resulting $M$ events and
their indices are reused across queries. Since each trajectory contains at most
$T_{\max}$ interaction steps, the segmentation and retrieval cost is:
\begin{equation}
\mathcal{C}^{\mathrm{I}}_{\mathrm{AiO}}
=
O(N\ell)\cdot\mathcal{C}_{\mathrm{LLM}}
+
O(M+T_{\max})\cdot\mathcal{C}_{\mathrm{emb}}
+
O(T_{\max}M)\cdot\mathcal{C}_{\mathrm{sim}} .
\end{equation}

On top of this, AiO-Mem keeps $\mathcal{B}$ unchanged and constructs only
the query-specific working memory $\mathcal{W}_t$. Each interaction step
operates on at most $K_{\mathrm{ret}}$ retrieved events with an average
exposed length of $\ell_e$ tokens. Therefore, the construction
cost is:
\begin{equation}
\mathcal{C}^{\mathrm{II}}_{\mathrm{AiO}}
=
O(T_{\max}K_{\mathrm{ret}}\ell_e)
\cdot\mathcal{C}_{\mathrm{LLM}} .
\end{equation}

Combining event-bank construction, multi-view retrieval, bounded
working-memory construction, and final response generation, AiO-Mem requires
\begin{equation}
\begin{aligned}
\mathcal{C}^{\mathrm{infer}}_{\mathrm{AiO}}
={}&
O(N\ell+T_{\max}K_{\mathrm{ret}}\ell_e)
\cdot\mathcal{C}_{\mathrm{LLM}}
\\
&+
O(M+T_{\max})\cdot\mathcal{C}_{\mathrm{emb}}
+
O(T_{\max}M)\cdot\mathcal{C}_{\mathrm{sim}} .
\end{aligned}
\end{equation}

\begin{figure*}[t]
    \centering

    \begin{subfigure}[b]{0.55\textwidth}
        \centering
        \includegraphics[
            height=4.8cm,
            keepaspectratio
        ]{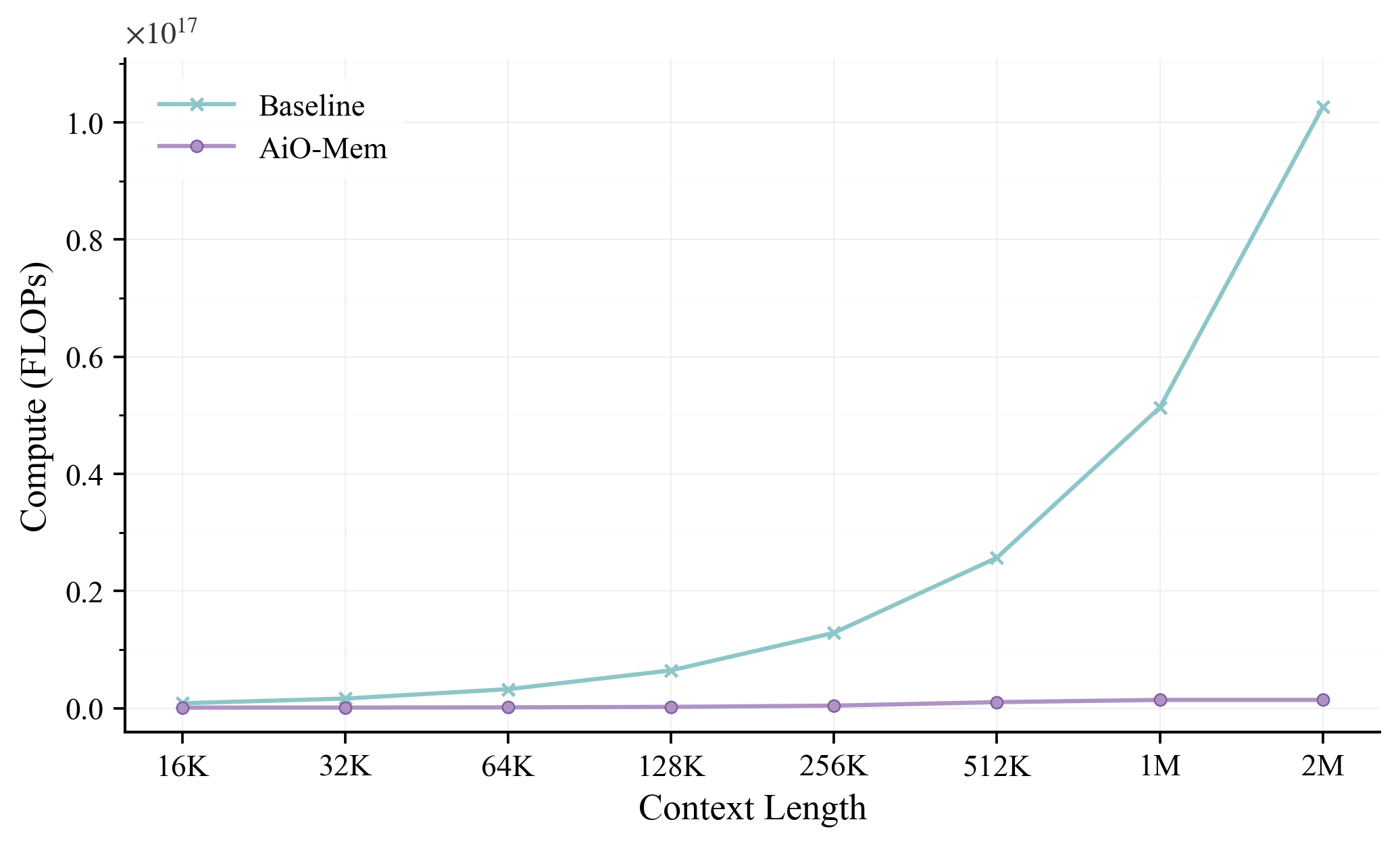}
        \caption{FLOPs comparison on LoCoMo.}
        \label{fig:efficiency-flops}
    \end{subfigure}
    \hfill
    \begin{subfigure}[b]{0.4\textwidth}
        \centering
        \includegraphics[
            height=4.5cm,
            keepaspectratio
        ]{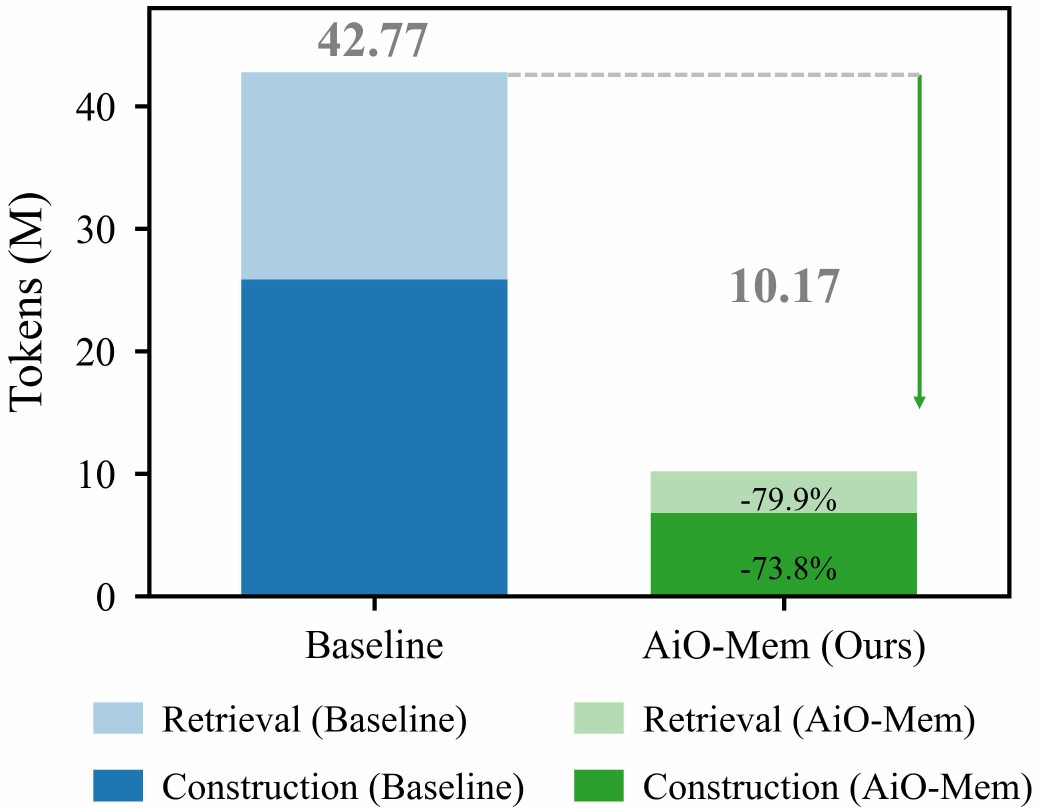}
        \caption{Token comparison on LoCoMo.}
        \label{fig:efficiency-token}
    \end{subfigure}

    \caption{Efficiency comparison between AiO-Mem and Baseline on LoCoMo.}
    \label{fig:efficiency-analysis}
\end{figure*}

Following prior efficiency-oriented works, we report FLOPs measured by \citep{calflops} and token consumption, using Mem-T as the baseline. As shown in Figure~\ref{fig:efficiency-analysis}, AiO-Mem exhibits substantially slower FLOPs growth with increasing context length. By decoupling large-scale persistent memory management from query-time working-memory construction, AiO-Mem avoids repeated memory rewriting and re-indexing and confines costly LLM computation to selected evidence. It further reduces retrieval and construction tokens by 79.9\% and 73.8\%, respectively.

\begin{table*}[t]
    \centering
    \caption{Notation used in the complexity analysis.}
    \label{tab:aio-complexity-notation}

    \small
    \setlength{\tabcolsep}{5pt}

    \begin{tabular}{@{}lp{0.75\linewidth}@{}}
    \toprule
    \textbf{Symbol} & \textbf{Definition} \\
    \midrule

    $\mathcal{H}=(h_1,\ldots,h_N)$
    & Long-horizon history consisting of $N$ heterogeneous source units. \\

    $N$
    & Number of source units in $\mathcal{H}$. \\

    $\ell$
    & Average number of tokens in each source unit $h_n$. \\

    $\mathcal{B}=\mathcal{C}(\mathcal{H})$
    & Traceable event bank constructed from $\mathcal{H}$ by the structure-aware compiler $\mathcal{C}$. \\

    $M$
    & Number of traceable events in the event bank $\mathcal{B}$. \\

    $M_1$
    & Number of persistent memory entries produced per source unit by OMS. \\

    $R_1$
    & Fraction of OMS memory entries that trigger memory rewriting and re-indexing. \\

    $K_{\mathrm{ret}}$
    & Number of candidate events returned by each \textsc{search} action. \\

    $\ell_e$
    & Average number of tokens exposed by one retrieved event. \\

    $T_{\max}$
    & Maximum interaction horizon, with $T\leq T_{\max}$. \\

    $\mathcal{C}_{\mathrm{seg}}$
    & Cost of one segmentation operation. \\

    $\mathcal{C}_{\mathrm{LLM}}$
    & Cost of LLM computation. \\

    $\mathcal{C}_{\mathrm{emb}}$
    & Cost of an embedding operation. \\

    $\mathcal{C}_{\mathrm{sim}}$
    & Cost of a sparse or dense similarity operation. \\

    \bottomrule
    \end{tabular}
\end{table*}

\clearpage

\section{Prompts and Tool Calls}
\label{app:prompts}

\subsection{Memory Agent Tool Calls}

The LoCoMo memory agent alternates between DECIDE and EXTRACT. DECIDE chooses between search and stop; after each search, EXTRACT chooses between add, which writes visible evidence into working memory, and skip.Table~\ref{tab:memory_tool_calls} lists the tools available at each stage.

\begin{table}[H]
\centering
\small
\setlength{\tabcolsep}{3pt}
\renewcommand{\arraystretch}{1.0}
\setlength{\abovecaptionskip}{4pt}
\setlength{\belowcaptionskip}{3pt}
\caption{Tool calls available to the memory agent.}
\begin{tabular}{@{}>{\raggedright\arraybackslash}p{1.25cm} >{\raggedright\arraybackslash}p{\dimexpr\linewidth-1.25cm-2\tabcolsep\relax}@{}}
\toprule
\textbf{Stage} & \textbf{Tool call and function} \\
\midrule
DECIDE
& \texttt{search\_memory\_units(query)}: Retrieves event-level memory units concerning information missing from the current notes. \\
DECIDE
& \texttt{answer\_from\_notes()} (no arguments): Ends memory retrieval and proceeds to the answering stage using the existing working memory. \\
EXTRACT
& \texttt{add\_note(event\_id, datetime, content, evidence\_turns)}: Saves a clue, its event time, and the original dialogue supporting it. An EXTRACT step may invoke \texttt{add\_note} multiple times, with each invocation writing one note. \\
EXTRACT
& \texttt{skip\_extraction(reason)}: Skips extraction when the current search results contain no relevant evidence to write. \\
\bottomrule
\end{tabular}
\label{tab:memory_tool_calls}
\end{table}

\subsection{LoCoMo Memory Agent Prompts}

Table~\ref{tab:memory_agent_prompts} presents condensed prompt templates for the two model-call stages. Each call includes only the tool definitions permitted at that stage; the model returns tool calls as JSON objects enclosed in \texttt{<tool\_call>} tags.

\begin{table}[H]
\centering
\small
\setlength{\tabcolsep}{3pt}
\renewcommand{\arraystretch}{1.0}
\setlength{\abovecaptionskip}{4pt}
\setlength{\belowcaptionskip}{3pt}
\caption{Prompt templates for the LoCoMo memory agent.}
\begin{tabular}{@{}>{\raggedright\arraybackslash}p{1.55cm} >{\raggedright\arraybackslash}p{\dimexpr\linewidth-1.55cm-2\tabcolsep\relax}@{}}
\toprule
\textbf{Stage} & \textbf{Prompt template} \\
\midrule
DECIDE
& You are a memory assistant. Given the current question and saved working memory, choose the next action: search memory, or stop retrieval and proceed to the answering stage.\newline
\textbf{Rules:}
(1) Judge whether the question can be answered completely using only the working memory, including the relationships and all necessary parts requested by the question. If the evidence is sufficient, call \texttt{answer\_from\_notes}.
(2) Otherwise, generate a query targeting one specific fact or clue that is still missing, and call \texttt{search\_memory\_units}. The query should identify the relevant person or object, relationship, and information needed. Avoid repeating previous searches that produced no useful notes.
(3) At the final decision step, call \texttt{answer\_from\_notes}.\newline
\textbf{Input:} Question \{question\}, working memory \{working\_notes\}, previous queries \{previous\_searches\}, and step \{step\}/\{max\_steps\}.\newline
\textbf{Output:} A brief explanation of the decision and one \texttt{<tool\_call>}. \\
\midrule
EXTRACT
& You are a memory system analyst. Given the question, the latest search results, and the existing working memory, identify new evidence useful for answering the question and write it into the working memory. Search results are temporary; useful information not written into the notes will not be available to subsequent decisions.\newline
\textbf{Rules:}
(1) Read the visible original dialogue in retrieved events. Highlighted turns are only hints; relevant evidence may also appear in other turns.
(2) For each new useful clue, call \texttt{add\_note} to write a concise, faithful note, preserving the exact event ID, recorded time, and original dialogue supporting its content. Each note should capture one natural clue and should be compared with existing notes to avoid duplication.
(3) If the search results contain no relevant evidence, call \texttt{skip\_extraction}.\newline
\textbf{Input:} Question \{question\}, search observation \{search\_observation\}, and working memory \{working\_notes\}. The search observation contains event identifiers, recorded times, retrieved fields, summaries, and visible original dialogue.\newline
\textbf{Output:} A brief assessment of the evidence and the required \texttt{<tool\_call>}; distinct clues may be written as separate notes. \\
\bottomrule
\end{tabular}
\label{tab:memory_agent_prompts}
\end{table}

\clearpage
\subsection{Stepwise Reward Model Prompt}

For SFT, the prompt below is the reward model input, with rule-derived scores as targets. For DPO, the same prompt is given to both the reward model and DeepSeek-V4-Flash; DeepSeek's score sequence serves as chosen and the rule-derived sequence as rejected. Reference answers and annotated evidence are used only for reward scoring, not by the memory policy or answer model. Table~\ref{tab:step_rm_prompt} summarizes the prompt's objectives, error categories, score ranges, and output format.

\begin{table}[H]
\centering
\small
\setlength{\tabcolsep}{3pt}
\renewcommand{\arraystretch}{1.0}
\setlength{\abovecaptionskip}{4pt}
\setlength{\belowcaptionskip}{3pt}
\caption{Prompt template and scoring rules for the stepwise reward model.}
\begin{tabular}{@{}>{\raggedright\arraybackslash}p{\linewidth}@{}}
\toprule
\textbf{Prompt template} \\
\midrule
\textbf{Task and input}\newline
You are a reward model evaluator of memory trajectories. Given a complete scoreable memory trajectory, evaluate each step according to whether the required evidence was found, whether useful evidence was written into working memory, and whether retrieval stopped when the evidence was sufficient. You may use trajectory context to understand the current action, but each score must evaluate only the current step; do not evaluate the quality of the final answer model's response.\newline
\textbf{Input:} \{question\}, \{reference\_answer\}, \{gold\_evidence\}, \{trajectory\_information\}, and \{scoreable\_steps\}.\newline
\textbf{Objectives by action type}\newline
\texttt{decide\_search}: Does the query target information missing from the current notes and help locate evidence needed to answer the question? A reasonable query should not receive the lowest score solely because it failed to retrieve a match.\newline
\texttt{extract\_add\_note}: Does the action faithfully write useful, previously unsaved evidence from the latest search into the notes, and do the cited turns support the note's content?\newline
\texttt{extract\_skip}: Does the latest search truly contain nothing worth writing? Skipping clearly critical evidence is a severe problem.\newline
\texttt{decide\_answer\_from\_notes}: Are the existing notes sufficient to support an answer when retrieval stops?\newline
\textbf{Evidence assessment}\newline
When judging whether the notes are sufficient, consider both their content and the original dialogue they cite. Annotated evidence indicates information needed for the question, but it is not a mechanical hit-rate scoring formula: unannotated dialogue turns may also support the answer semantically, and matching annotated turns does not automatically warrant a high score.\newline
\textbf{Error severity}\newline
\textbf{Category A, high severity:} Failure of the current stage's core objective.\newline
\textbf{Category B, high severity:} Factual or attribution errors in a note, or citations that do not support key content.\newline
\textbf{Category C, medium severity:} Problems in assessing the current state or respecting stage boundaries.\newline
\textbf{Category D, medium severity:} An incomplete contribution or substantial noise in the action.\newline
\textbf{Category E, low severity:} Wording or formatting issues.\newline
\textbf{Score ranges}\newline
\textbf{90--100:} The action fulfills the stage's core objective very well, with at most minor Category E issues.\newline
\textbf{75--89:} The action is clearly useful and largely fulfills the core objective; it has no Category A or B issues, though minor Category C, D, or E issues are allowed.\newline
\textbf{60--74:} The action makes a partial contribution but falls short of the core objective; it has no Category A or B issues, though Category C or D issues are more pronounced.\newline
\textbf{40--59:} The action has a severe problem but retains some useful contribution; it typically contains one Category A or B issue.\newline
\textbf{1--39:} The action provides almost no useful contribution or takes a clearly wrong direction; severe problems cause the current step to largely fail.\newline
\textbf{Output requirements}\newline
Output a strict JSON object containing \texttt{step\_scores} in input order. Each step must contain \texttt{step\_index}, \texttt{step\_type}, \texttt{quality\_level}, and an integer \texttt{score} from 1 to 100. Do not omit or add steps, and do not output scoring explanations.\newline
\texttt{quality\_level} roughly corresponds to the score as follows: \texttt{bad} (1--30), \texttt{weak} (31--55), \texttt{partial} (56--70), \texttt{good} (71--85), and \texttt{excellent} (86--100). \\
\bottomrule
\end{tabular}
\label{tab:step_rm_prompt}
\end{table}

\end{document}